\documentclass{article}
 \usepackage[preprint,nonatbib]{neurips_2021}

\usepackage[utf8]{inputenc} %
\usepackage[T1]{fontenc}    %
\usepackage{hyperref}       %
\usepackage{url}            %
\usepackage{booktabs}       %
\usepackage{amsfonts}       %
\usepackage{nicefrac}       %
\usepackage{microtype}      %

\usepackage{graphicx}
\usepackage{subfigure}
\usepackage[table,xcdraw]{xcolor}

\usepackage{tabulary}
\usepackage{colortbl}
\usepackage{diagbox}

\usepackage{color,soul}

\newcommand{\ourshort}{\texttt{SceneIT}}
\newcommand{\ourlong}{\texttt{Scene Images for Text-based Games}}
\newcommand{\tw}{\textit{TextWorld}}
\newcommand{\twc}{\textit{TWC}}
\newcommand{\twclong}{\textit{TextWorld Commonsense}}
\newcommand{\ftwp}{\textit{FTWP}}
\newcommand{\jericho}{\textit{Jericho}}

\definecolor{EasyColor}{rgb}{0.141, 0.388, 0.290}
\definecolor{MediumColor}{rgb}{0.588, 0.443, 0.090}
\definecolor{HardColor}{rgb}{0.702, 0.106, 0.106}   

\title{{\em Eye of the Beholder}:\\ Improved Relation Generalization for\\ Text-based Reinforcement Learning Agents}

\author{%
  Keerthiram Murugesan\\
  IBM Research\\
  \texttt{keerthiram.murugesan@ibm.com} \\
  \And
  Subhajit Chaudhury \\
  IBM Research\\
  \texttt{subhajit@jp.ibm.com} \\
  \And
  Kartik Talamadupula \\
  IBM Research\\
  \texttt{krtalamad@us.ibm.com} \\
}

\begin{document}

\maketitle

\begin{abstract}
  Text-based games (TBGs) have become a popular proving ground for the demonstration of learning-based agents that make decisions in quasi real-world settings. The crux of the problem for a reinforcement learning agent in such TBGs is identifying the objects in the world, and those objects' relations with that world. While the recent use of text-based resources for increasing an agent's knowledge and improving its generalization have shown promise, we posit in this paper that there is much yet to be learned from visual representations of these same worlds. Specifically, we propose to retrieve images that represent specific instances of text observations from the world and train our agents on such images. This improves the agent's overall understanding of the game {\em scene} and objects' relationships to the world around them, and the variety of visual representations on offer allow the agent to generate a better generalization of a relationship. We show that incorporating such images improves the performance of agents in various TBG settings.  
\end{abstract}

\section{Introduction}
\label{sec:introduction}

Reinforcement Learning (RL) has seen a resurgence in recent years thanks to advances in representation, inference, and learning techniques -- led by a massive scale-up and investment in deep neural network-based methods. Successful applications of RL have included domains such as Chess~\cite{silver2018general}, Go~\cite{silver2017mastering}, and Atari games~\cite{mnih2016asynchronous}. However, with the emergence of natural language processing (NLP) as a key AI application area, research attention has turned towards text-based applications and domains. These domains offer their complexity challenges for RL algorithms, including large and intractable action spaces -- the space of all possible words and combinations; partial observability of the world state; and under-specified goals and rewards. 

Text-based games (TBGs) have emerged as prime exemplars of the above challenges. Inspired by games such as Dungeons \& Dragons and Zork, researchers have worked on putting together challenging environments that offer the complexities of real-world interactions but in sandbox settings suitable for the training of RL agents. The foremost such example is \tw~\cite{cote2018textworld}, an open-source text-based game engine that allows for the generation of text-based game instances and the evaluation of agents on those games. Much of the recent work on text-based RL~\cite{ammanabrolu2019playing,dambekodi2020playing,murugesan2021text} has focused on the \tw\ environment, and on imbuing agents with additional information to make them learn, scale, and act more efficiently.

However, much of the information that has been used in the prior art to improve the performance of AI agents in TBGs is still restricted to the medium of text. In contrast, when humans encounter games such as Zork and \tw, they do not restrict themselves to only textual information. Indeed, they are able to generalize to environments and the actions within them by considering not just the form of information provided by the environment; but also by {\em imagining} or visualizing various forms of that information. This imagination is key to generalizing beyond merely the information present in the instance currently under consideration. In this work, we posit that using images -- either retrieved or imagined (generated) -- that represent information from the game instance can help improve the performance of RL agents in TBGs.

Specifically, we consider RL agents in the \tw\ and \jericho\ TBG environments; and additional information that can be provided to such agents to improve their performance. Past work has focused on trying to use external knowledge to either limit~\cite{chaudhury2020bootstrapped} or enhance~\cite{murugesan2021text} the space of actions: however, this has also been restricted to the text modality. At their crux, these efforts are all trying fundamentally to solve the problem of {\em relationships} within the environment -- {\em how are different things in the world related to each other?} And how can the agent manipulate these relations to convert the initial state of the world -- via a sequence of observations -- into the desired goal state (or to maximize reward)? Purely text-based information is extremely sparse and is unable to sufficiently abstract the notion of relationships.

Consider for example the relationship {\tt at} - a {\tt patio chair} is {\tt at} the {\tt backyard}. What does this relation mean - what is the {\em at}-ness? Text cannot convey this information effectively on its own: as the size of the underlying vocabulary increases, the natural language space gets sparser and it becomes harder to extract signals to understand relationships between objects (in this case, `patio chair' and `backyard'). Images, on the other hand, go a bit further in conveying the meanings of relationships as understood by humans~\cite{chen2015mind}. Images also help generalize better: in text, a patio chair is always represented as {\tt patio chair}; yet in a visual medium, there can exist different kinds of patio chairs, with different properties such as shape, size, color, texture, surroundings, etc. 

In this paper, we introduce the \textcolor{black}{\ourlong\ (\ourshort)} model (pronounced ``{\it seen-it}'') that integrates an external repository of images as additional knowledge for an RL agent in text-based game environments; and measure the performance of this model against the state-of-the-art text-only method. Our images come from two sources: pre-retrieved from prior existing images; and generated anew based on textual descriptions. We show that an agent with access to this additional visual information does significantly better, and examine some specific instances that show the reason for this improved performance.

\section{Methodology}
\label{sec:methodology}

Text-based reinforcement learning agents for TBGs interact with the environment only using the modality of text \cite{narasimhan2015language, he2016deep}. TBGs convey the state of the game at every step as observations in natural language text, and the text-based RL agent learns to map the current state to one of the admissible actions (also in the text modality) available to it. Most current text-based RL agents (e.g. \cite{murugesan2021text}) focus on integrating additional textual knowledge to learn and act in a complex environment. Such agents thus lack the ability for human-like imagination involved in solving TBGs efficiently.

In this section, we outline the methodology that we use to integrate the visual (image) representation of a game scene using our \ourshort\ approach for TBGs. In order to obtain the visual representation of the scene that the agent is currently situated in, as the first step, we extract noun phrases that represent objects and relational phrases between the objects in the scene from the text observation -- for example, \texttt{kitchen of the white house}, \texttt{bottle on the table}, \texttt{desk chair at bedroom}, etc. These phrases portray the scene in terms of which object is located at what location, which we intend to use to create a ``visual mind-map'' of the scene for the agent.

Since the key component and novelty of our system is the usage of images for the TBGs under consideration, we first outline the collection process for such images. Our technique relies on two main sources of images: {\em retrieval} from the internet, and {\em generation} from pre-existing models for imagining and generating visual scenes. We describe each of these methods in detail below.

\begin{figure*}[htb]
		\centering
		\includegraphics[width=0.99\linewidth]{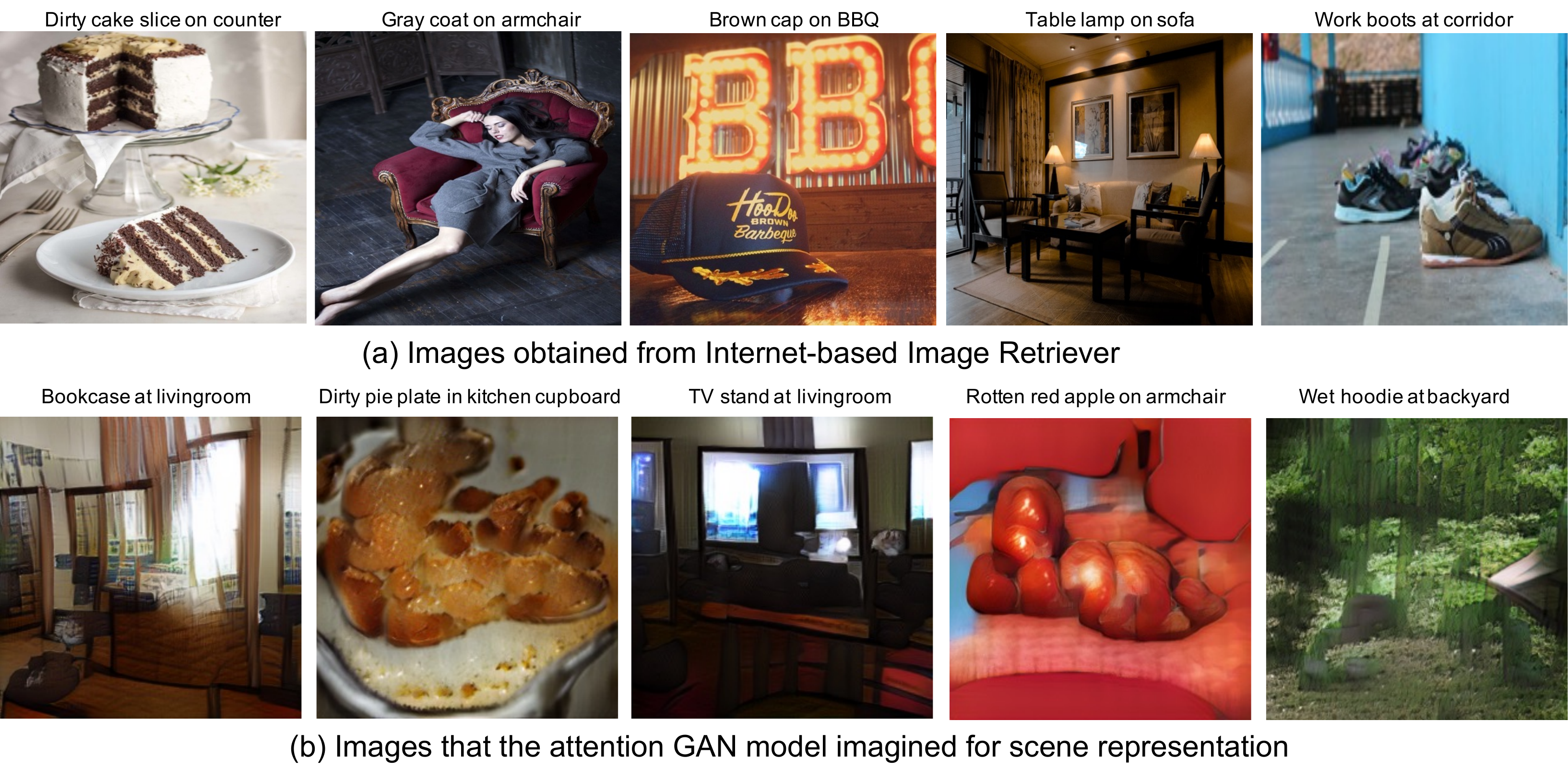}
		\caption{Examples of images obtained from (a) the web-based image retriever, and (b) imagination via AttnGAN~\cite{xu2018attngan}. The phrase used to retrieve or generate the picture is indicated above the respective picture.}
		\label{fig:retrieved_imagined_images}
\end{figure*}

\subsection{Collecting Images}
\label{subsec:collecting_images}
\textbf{Retrieving Images from the Internet:}
In order to obtain images from the internet, we design an image retriever that obtains the best matching image from the list of query strings (noun phrases) that are used to represent the scene. This process also ties into one of the central motivations of our work, which is that images offer more signals to agents as they try to abstract, represent, and use the relationships between different objects in a scene.

To provide good generalization behavior, we design an image retriever that automatically searches the internet for a given query string without any human supervision \footnote{Based on Google Image Retriever: \url{https://github.com/Joeclinton1/google-images-download}}. 
In addition, we use image caching to improve the speed of retrieval such that the images corresponding to encountered queries are saved to disk and need not be downloaded from the web while training the agent. It is to be noted that the caching process is completely generic and does not involve saving specific situation-relevant images. Figure~\ref{fig:retrieved_imagined_images}(a) provides some examples of images that are retrieved from the internet for specific phrases.

\textbf{Imagining Images from Generative Models:} The previous method of ``visual mind-map'' extraction uses pre-existing images from the internet for scene representation. Such a scene representation is useful for a human to visually parse the scene. However, we also explore the potential for representing visual scenes using images that are {\em imagined} by generative models. Our hypothesis is that such images can also provide useful visual features to improve generalization in tasks from \tw\ (and other text-based games).

We use the Attentional Generative Adversarial Network (AttnGAN)~\cite{xu2018attngan} for attention-driven text-to-image generation. This generative model uses a multi-stage refinement for fine-grained generation of images from a given text snippet. AttnGAN gives attention to the relevant tokens in the natural language query in order to generate details at different sub-regions of the image. 

For our approach, we pre-train the AttnGAN model on the MS-COCO dataset~\cite{lin2014microsoft}. The queries used for image generation are the same as the ones used for the previous internet retrieval-based scene representation. We hypothesize that although such images may not always be interpretable by humans -- see Figure~\ref{fig:retrieved_imagined_images}(b) for a few examples -- such images can provide some latent image features for neural models that might contribute to better generalization in \tw\ games.

\subsection{Model Description}
\label{subsec:model}
\label{sec:models}

\begin{figure*}
    \centering
    \includegraphics[width=1.0\linewidth]{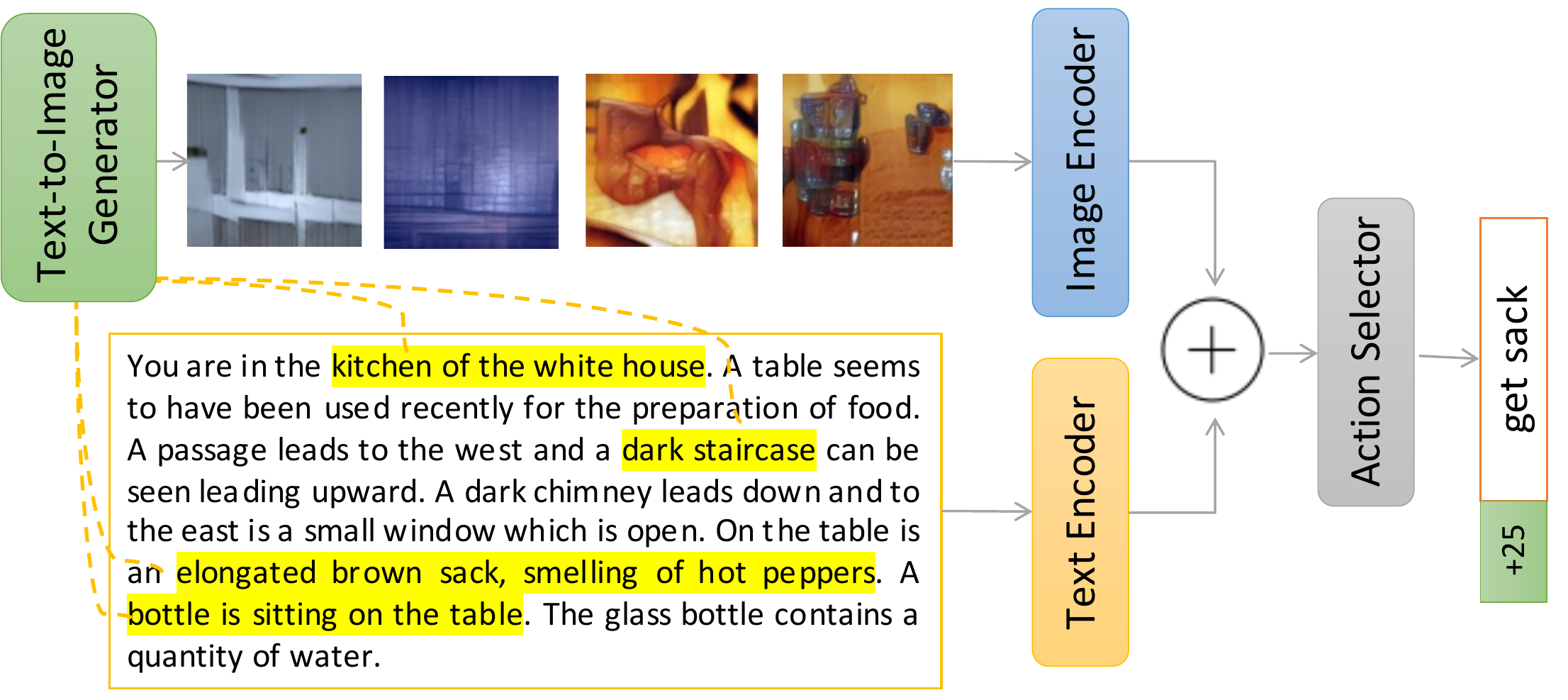}
    \caption{Overview of our methodology of scene representation for a sample text observation taken from \texttt{Zork1} using text-to-image generative model. \hl{Highlighted} text snippets show some of the phrases used by the agent to generate relevant images for scene representation.}
    \label{fig:arch_modelgan}
\end{figure*}

We now detail the models that we used to use and encode the images retrieved or generated in the previous step. Figure \ref{fig:arch_modelgan} shows the architecture overview of our proposed approach for scene representation using the AttnGAN~\cite{xu2018attngan} based text-to-image generation. 
In order to capture the textual features from the text observation from the game, we use Stacked GRU as our text encoder: this keeps tracks of the state of the game across time steps.
Once we have the images retrieved/generated from the text snippets, we extract the image features using image encoders which are combined with the features from textual inputs to obtain the action scores.

Specifically, we use Resnet-50 for encoding the retrieved images and for the images generated from the pre-trained AttnGAN. The text and image encoding features are then concatenated and passed to the action selector (as shown in Figure~\ref{fig:arch_modelgan}), which maps the encoding features to action scores using a multi-layer perceptron (MLP) to select the next action. Based on the reward from the game environment, we update text and image encoders and the action selector. Since the reward from the game can guide the text-to-image generator (AttnGAN) to generate meaningful images for the current context of the game, we finetune the pre-trained AttnGAN along with the encoders and the action selector to yield the best results. In this case, we use the inbuilt CNN-based image encoder ({Inception~v3}~\cite{szegedy2016rethinking}) to map the generated images to the image features. We call this model \ourshort\ and use it by default for all our experiments in this paper.

\section{Experimental Results}
\label{sec:results}

In this section, we present experimental results that demonstrate the advantage of our proposed \ourlong\ approach -- which makes use of images in addition to text -- over existing state-of-the-art techniques that are text-only. We conduct our performance evaluation on three datasets: TextWorld Commonsense (\twc) \footnote{\url{https://github.com/IBM/commonsense-rl}}, the First TextWorld Problems (\ftwp) \footnote{\url{https://competitions.codalab.org/competitions/21557}} and \jericho \footnote{\url{https://github.com/microsoft/jericho}}. The \twc\ and \ftwp\ datasets build on the Microsoft \tw\ Environment \cite{cote2018textworld}, and offer complementary tests: while {\twc} tasks require the retrieval and use of commonsense knowledge for more efficient solution, the {\ftwp} problems test the agent's exploration capabilities. \jericho\ is a suite of $33$ interactive fiction games that measures human performance on text-based games by offering stories from different domains -- in our case, it helps evaluate the breadth and coverage of the image generation.

{\bf Distribution:} In these datasets, a set of text-adventure games are provided for training reinforcement learning (RL) agents. In addition to these training games, the datasets contain two test sets of games: 1) Test games (IN) that are generated from the same distribution as the training games -- these games contain similar sets of entities and relations as the train games; and 2) Test games (OUT), which contain games generated from a set of entities that have no overlap with the training games. This is a way of testing whether the RL agent can generalize its behavior to new and unseen games by leveraging the state observation from the TextWorld environment -- and additionally in our case, the visual relationships between entities.

{\bf Agents:} We compare three RL agents in our experiments: 1) \textit{Random}, where the actions are selected randomly at each step; 2) {\it Text-Only}, where the actions are selected solely based on the textual observation available at the current step. We use three baseline text-only methods - DRRN~\cite{he2016deep}, Template DQN~\cite{hausknecht19} and KG-A2C~\cite{ammanabrolu2020graph}; and 3) Our method -- \ourshort\ -- explained in the previous section, where the RL agent is allowed to imagine visual scenes and images using Attention GAN \cite{Tao18attngan}, a Text-to-Image generator based on Generative Adversarial Networks (GAN) \cite{NIPS2014_5ca3e9b1}.

{\bf Metrics:} In our experiments, we measure the performance of various agents using two metrics: (1) {\em Average Normalized Score} -- calculated as the total score achieved by an agent normalized by the maximum possible score for the game); and (2) {\em Average Steps Taken} -- calculated as the total number of steps taken by the agent to complete the goals. A higher score is better, while a lower number of steps taken is better.

\begin{figure*}[t]
    \centering
        \includegraphics[trim={0.6cm 0.5cm 0.25cm 0.2cm},scale=0.29]{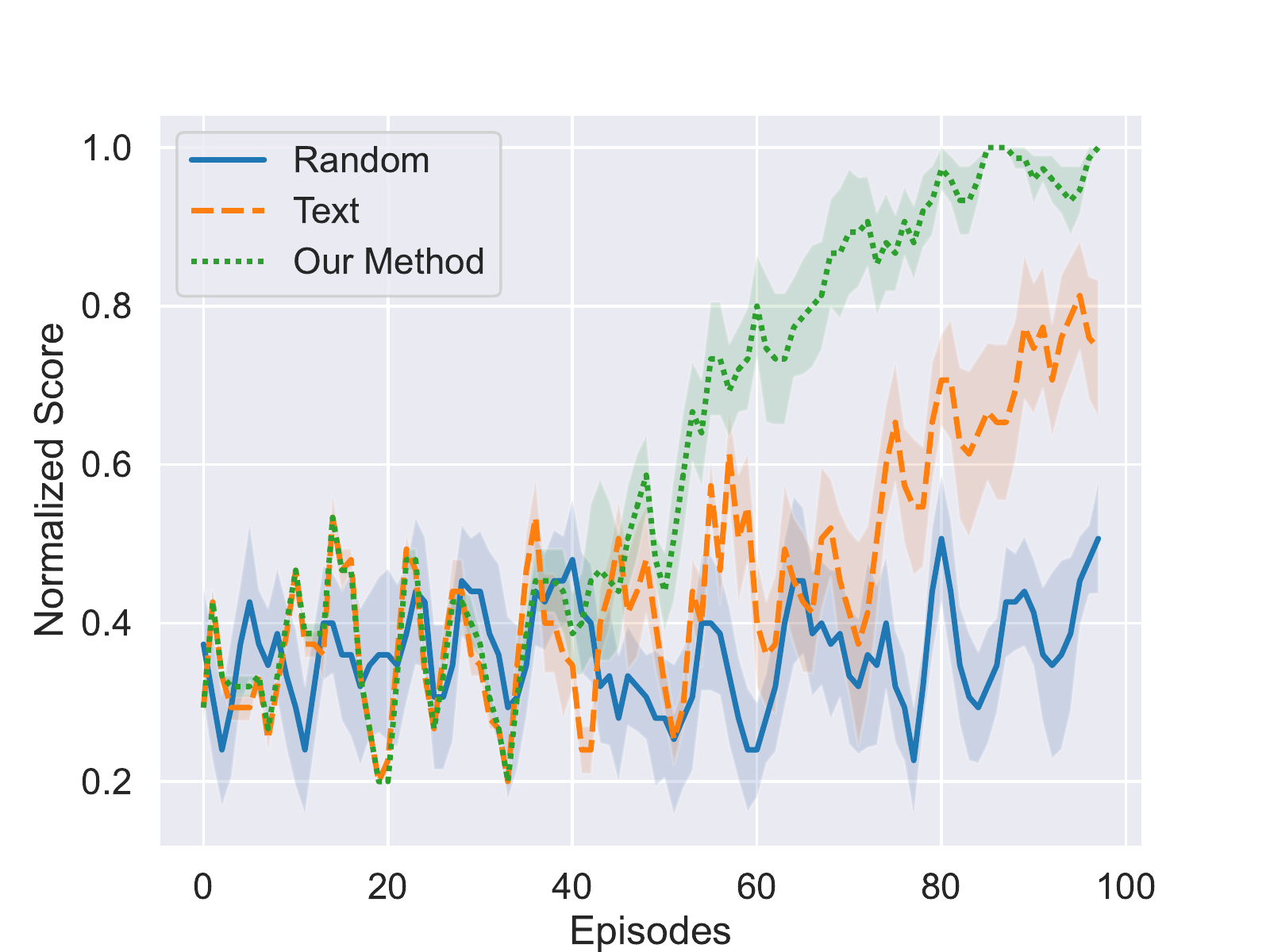} 
        \includegraphics[trim={0.6cm 0.5cm 0.25cm 0.2cm},scale=0.29]{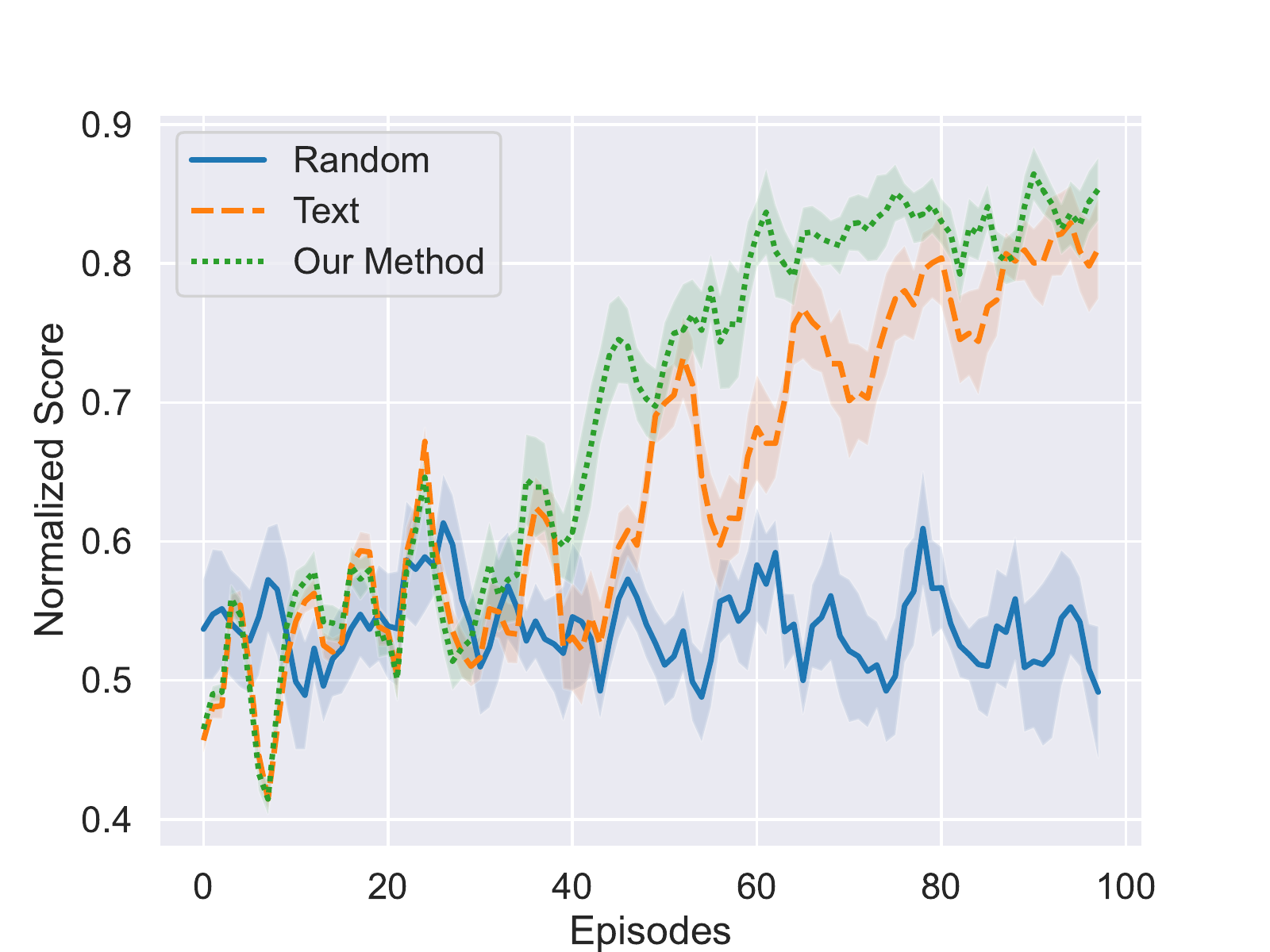}
        \includegraphics[trim={0.6cm 0.5cm 0.25cm 0.2cm},scale=0.29]{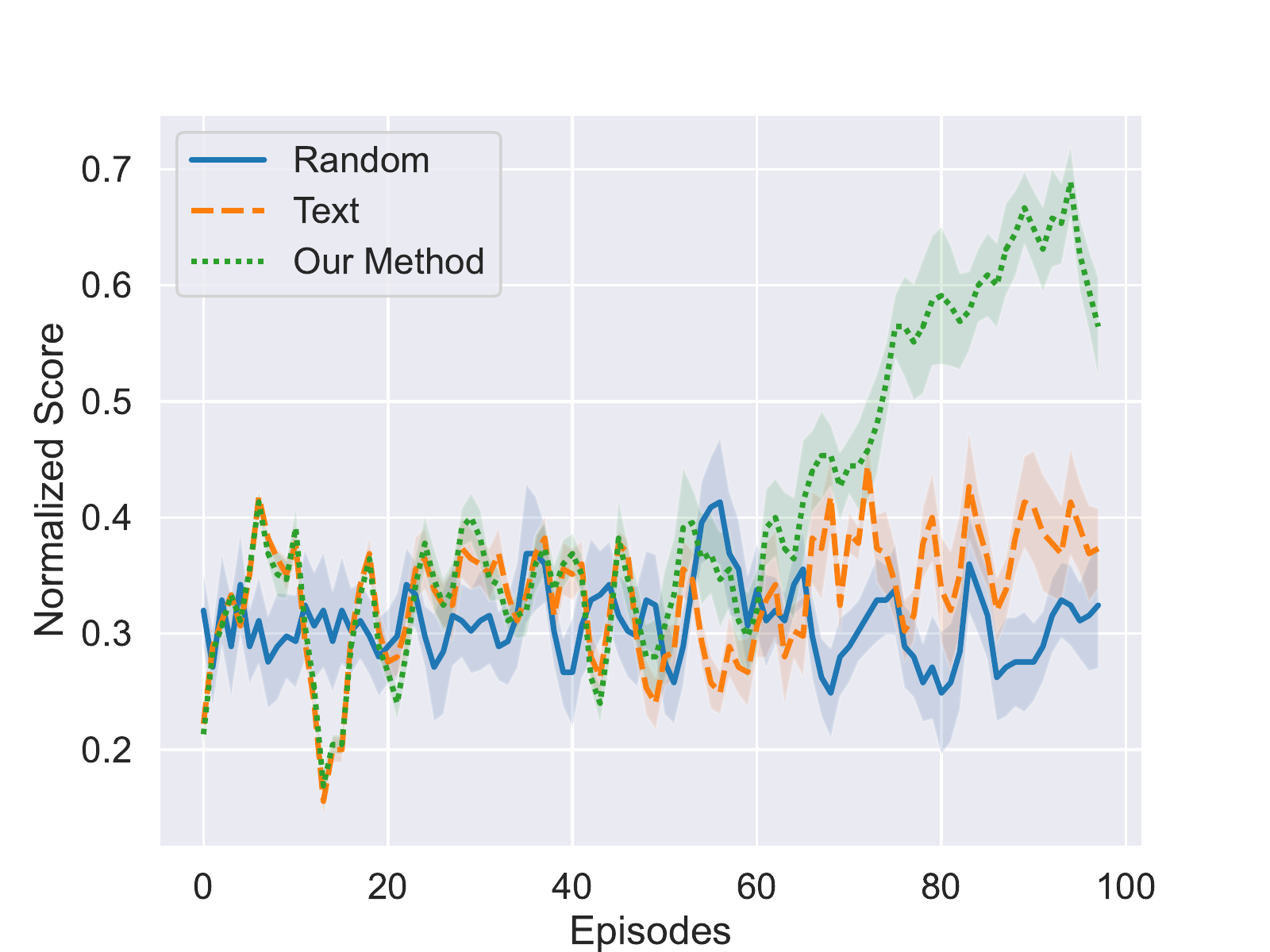}\\
        \includegraphics[trim={0.6cm 0.5cm 0.25cm 0.2cm},scale=0.29]{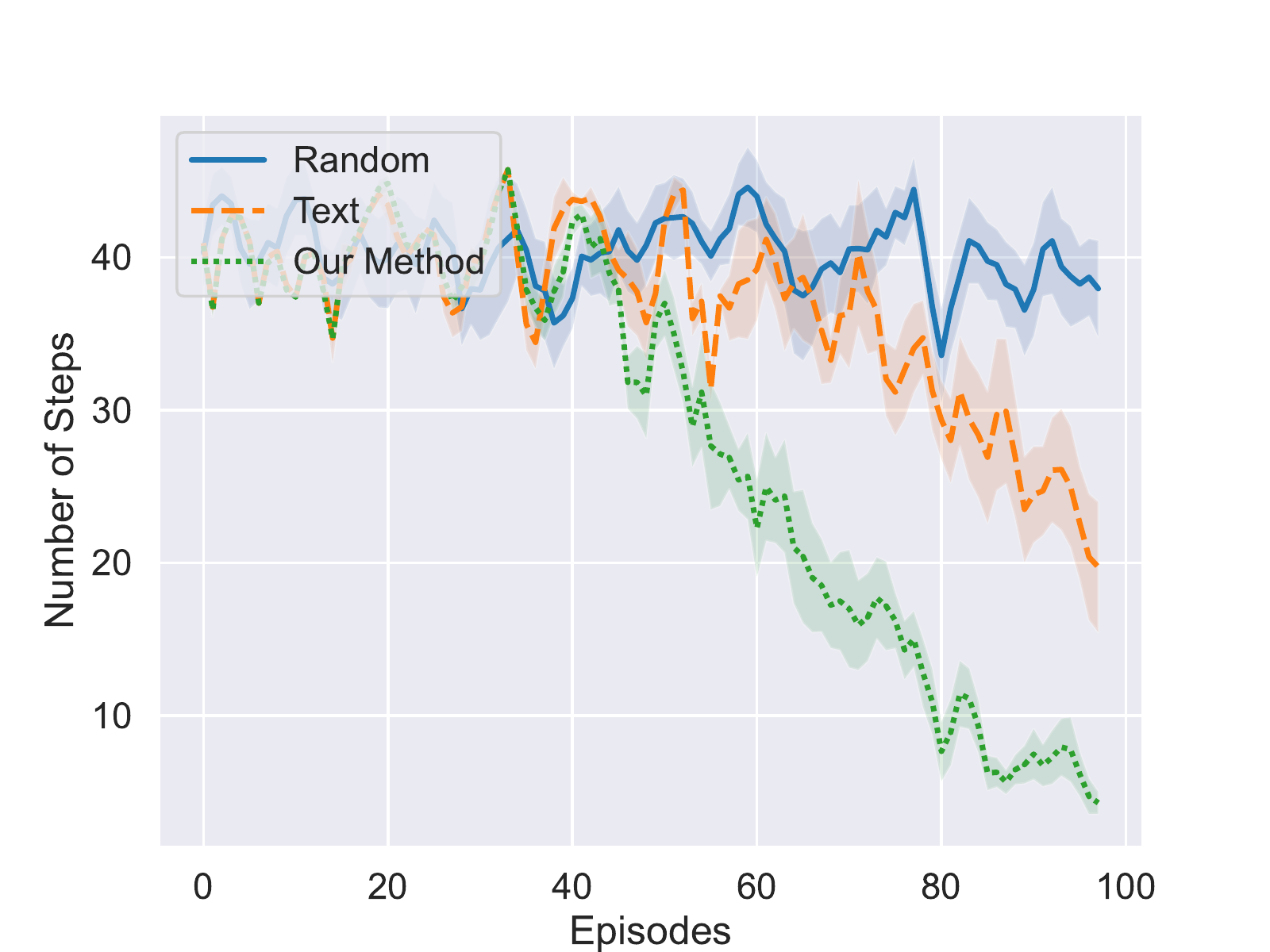}
        \includegraphics[trim={0.6cm 0.5cm 0.25cm 0.2cm},scale=0.29]{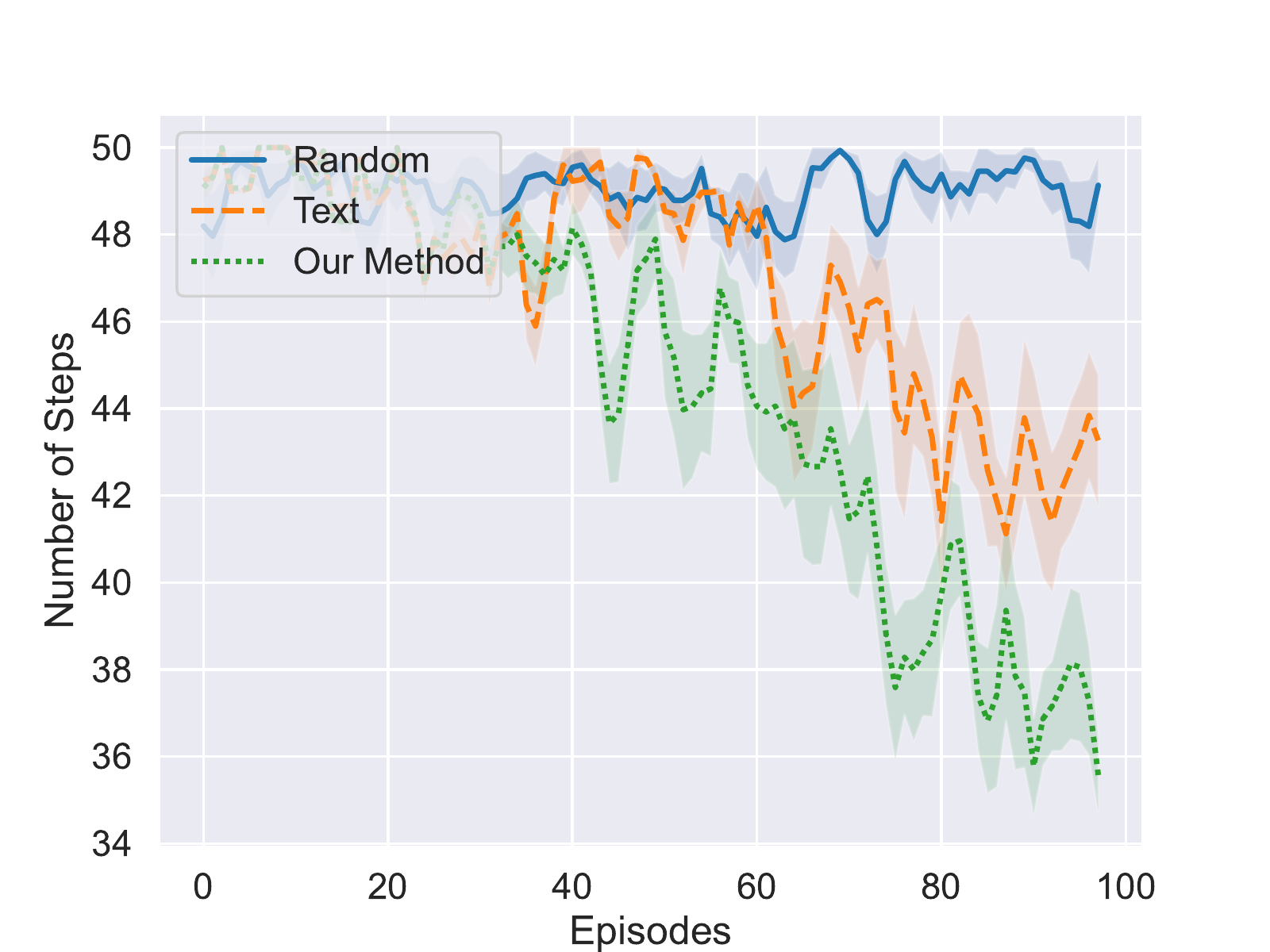}
        \includegraphics[trim={0.6cm 0.5cm 0.25cm  0.2cm},scale=0.29]{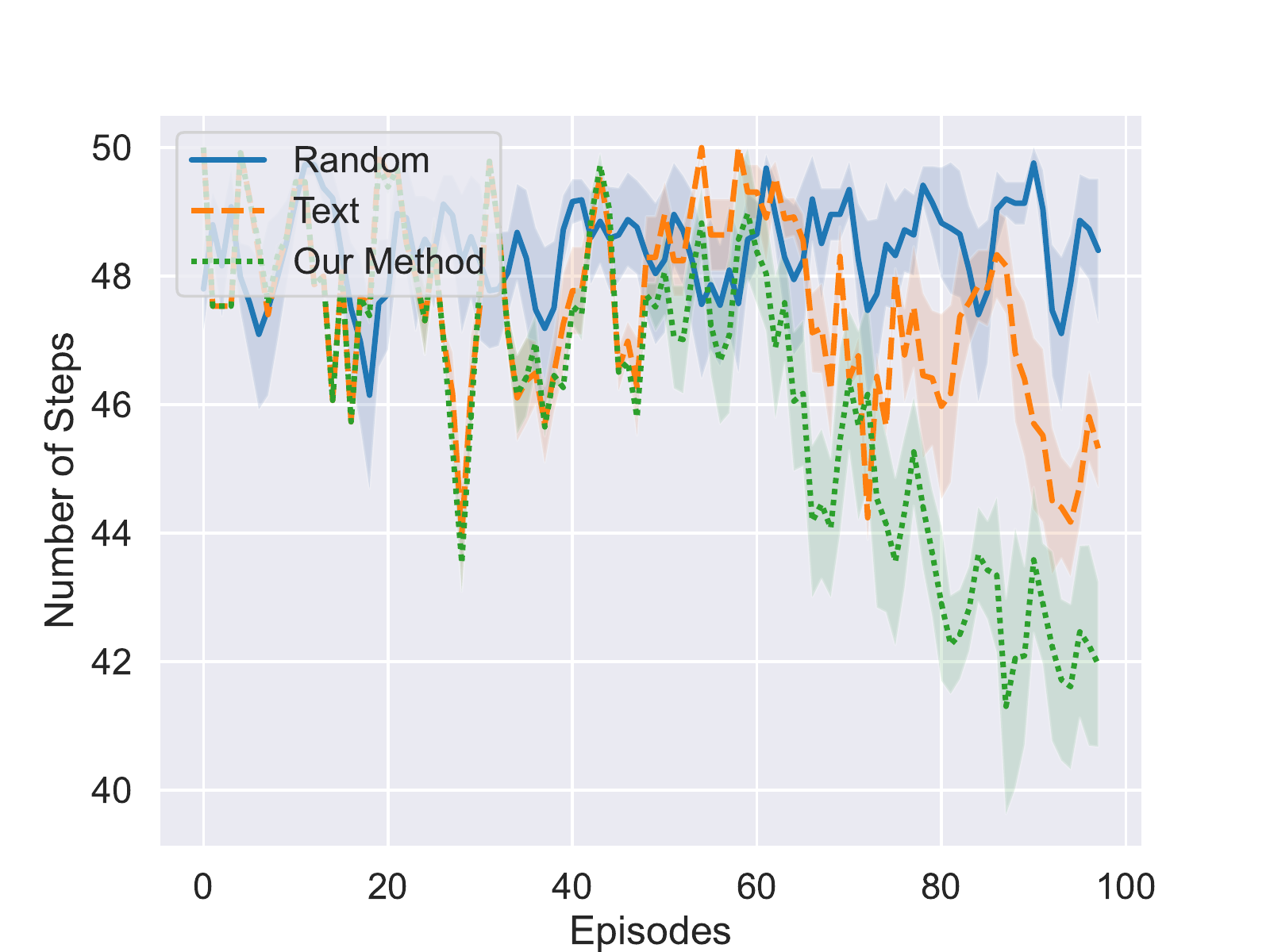}\\
    \caption{Training performance (showing mean and standard deviation averaged over \textcolor{black}{$5$} runs) for the three difficulty levels: Easy (left), Medium (middle), Hard (right). Higher normalized score is better, while lower number of steps is better. {\tt Our Method} refers to our \ourshort\ technique.}
    \label{fig:train_performance_twc}
\end{figure*}

\subsection{Quantitative Results}
\label{subsec:quantitative_results}

We first present the results of a quantitative evaluation of our proposed technique. In order to provide a well-rounded evaluation, we consider different text-based games: the \twc\ and \ftwp\ problems, both based on the \tw~\cite{cote2018textworld} domain; and the \jericho~\cite{hausknecht19} domain, based on interactive fiction (IF) games. Detailed experimental setting are provided in the supplementary material.

\subsubsection{Experiments on \twclong}

The first domain that we conduct our evaluation on is the \twclong~\cite{murugesan2021text} domain. This domain is an extension of the \tw\ domain that adds scenarios where commonsense knowledge is required in order to arrive at efficient solutions.

{\bf Difficulty Levels:} The \twc\ domain comes with difficulty levels for the problem instances associated with it, defined in terms of how hard it is for an agent (human or AI) to solve that specific instance. The difficulty of a level is set as a combination of the number of goals to be achieved, the number of actions (steps) required to achieve them, and the number of objects and rooms in the instance (which may be related to goal achievement, or may simply be distractors). In our evaluation for this work, we consider three distinct difficulty settings. In increasing order of hardness, these are: {\tt easy}, {\tt medium}, and {\tt hard}. 
We follow Murugesan {\it et al.}~\cite{murugesan2021text} -- who introduce the \twc\ domain, and are the current state-of-the-art on this domain -- in choosing these difficulty levels.

{\bf Training Performance:} Figure~\ref{fig:train_performance_twc} shows the training performance of three different agents/models on the \twc\ problems for the three difficulty levels discussed above. For each level, the performance is reported via the normalized score (higher is better) as well as the average number of steps (lower is better). It is clear that \ourshort\ -- with access to both the textual representation of the observations from the game, as well as the image/visual representation -- does much better in all three settings. Furthermore, beyond the $60$ episode mark, there is a clear divergence of our technique from the random and text-only baselines.

\begin{table*}[t]
\centering
\begin{tabular}{l|l|l|l|l|l|l|l|}
\cline{2-8}
\texttt{Norm. Score} &
  \multicolumn{3}{c|}{\cellcolor[HTML]{DAE8FC}Test Games (IN)} &
  \multicolumn{1}{c|}{\cellcolor[HTML]{DAE8FC}} &
  \multicolumn{3}{c|}{\cellcolor[HTML]{5885B6}Test Games (OUT)} \\ \hline
\multicolumn{1}{|l|}{\diagbox{Model}{Level}} &
  \cellcolor[HTML]{EFEFEF}Easy &
  \cellcolor[HTML]{EFEFEF}Medium &
  \cellcolor[HTML]{EFEFEF}Hard &
   &
  \cellcolor[HTML]{EFEFEF}Easy &
  \cellcolor[HTML]{EFEFEF}Medium &
  \cellcolor[HTML]{EFEFEF}Hard \\ \cline{1-4} \cline{6-8} 
\multicolumn{1}{|l|}{\cellcolor[HTML]{6A98C3}Random}   & 0.52           & 0.49         & 0.49      &  & 0.51           & 0.54           & 0.31      \\ \cline{1-4} \cline{6-8} 
\multicolumn{1}{|l|}{\cellcolor[HTML]{F19D60}Text}   & 0.82           & \textbf{0.74}         & 0.62      &  & 0.75           & 0.69           & 0.41      \\ \cline{1-4} \cline{6-8} 
\multicolumn{1}{|l|}{\cellcolor[HTML]{579D2E}\ourshort}   & \textbf{0.96}           & 0.70         & \textbf{0.77}      &  & \textbf{0.88}           & \textbf{0.78}           & \textbf{0.59}      \\ \hline 
\end{tabular}
\bigskip\\
\begin{tabular}{l|l|l|l|l|l|l|l|}
\cline{2-8}
\texttt{Num. Steps} &
  \multicolumn{3}{c|}{\cellcolor[HTML]{DAE8FC}Test Games (IN)} &
  \multicolumn{1}{c|}{\cellcolor[HTML]{DAE8FC}} &
  \multicolumn{3}{c|}{\cellcolor[HTML]{5885B6}Test Games (OUT)} \\ \hline
\multicolumn{1}{|l|}{\diagbox{Model}{Level}} &
  \cellcolor[HTML]{EFEFEF}Easy &
  \cellcolor[HTML]{EFEFEF}Medium &
  \cellcolor[HTML]{EFEFEF}Hard &
   &
  \cellcolor[HTML]{EFEFEF}Easy &
  \cellcolor[HTML]{EFEFEF}Medium &
  \cellcolor[HTML]{EFEFEF}Hard \\ \cline{1-4} \cline{6-8} 
\multicolumn{1}{|l|}{\cellcolor[HTML]{6A98C3}Random}   & 38.52           & 49.66         & 46.21      &  & 38.92           & 48.94           & 48.95      \\ \cline{1-4} \cline{6-8} 
\multicolumn{1}{|l|}{\cellcolor[HTML]{F19D60}Text}   & 22.73           & 46.36         & 39.54      &  & 30.18           & 46.29           & 46.90      \\ \cline{1-4} \cline{6-8} 
\multicolumn{1}{|l|}{\cellcolor[HTML]{579D2E}\ourshort}   & \textbf{13.38}          & \textbf{46.15}         & \textbf{34.65}      &  & \textbf{19.58}           & \textbf{38.18}           & \textbf{44.08}      \\ \hline 
\end{tabular}
\caption{Test performance (averaged over $5$ runs) on the normalized score (higher is better) and number of steps (lower is better) metrics for the three difficulty levels.}
\label{tab:test_results_full}
\end{table*}

{\bf Test Performance:} Table~\ref{tab:test_results_full} shows the test results for $3$ models - one random baseline, one text-only baseline, and \ourshort\ -- which combines the text features with image features from the finetuned AttnGAN. We split our reporting across two conditions: Test games (IN) reports on test games that come from the same distribution as the training games; while Test games (OUT) reports on test games from outside the distribution of training games. It is clear that for both conditions, \ourshort\ is the state-of-the-art in $11$ out of $12$ instances -- handily beating the existing text-only state-of-the-art ({\tt Text}). In the one case where it is not the best ({\tt medium} for {\tt in} distribution), it is very close to the performance of the best performing model. This shows the added advantage of using visual features in addition to textual features when solving \twc\ games, thus validating the central hypothesis of our work.

\subsubsection{Experiments on \textit{First TextWorld Problems}}

\begin{figure*}[htbp]
    \centering
    \includegraphics[width=0.95\textwidth]{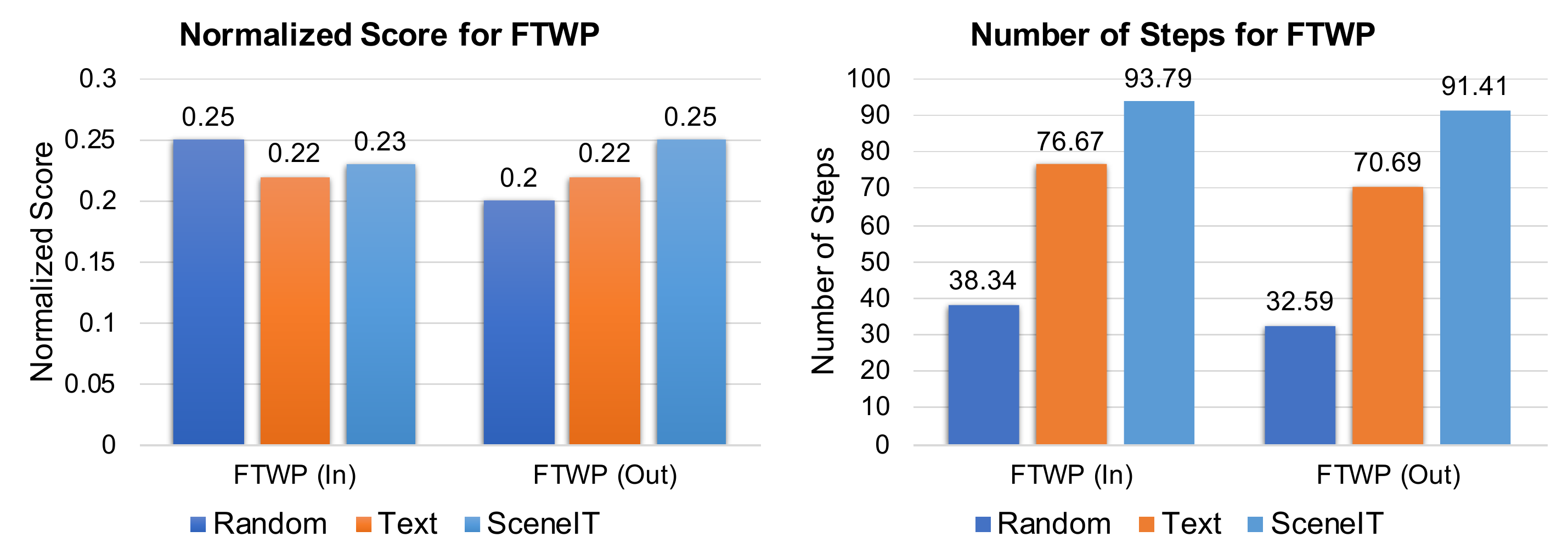}
    \caption{Test-set performance on FTWP Cooking Task (averaged over $5$ runs) on the normalized score (higher is better) and the number of steps (lower is better) metrics.}
    \label{fig:results_ftwp}
\end{figure*}

In this section, we present the results of running the various agents/models on the {\it First TextWorld Problems} (\ftwp) dataset. Figure~\ref{fig:results_ftwp} shows the results across the {\tt in} and {\tt out} distributions, as introduced previously. Since the cooking task in FTWP focuses more on exploration rather than the meaningful relationship between the objects (as in \twc\ ) to improve the performance,
we can see that \ourshort\ shows results that are comparable to and even worse than the text-only model: this shows that merely adding images to a game does not always necessarily improve the metrics.

\subsubsection{Experiments on \textit{Jericho}}

\begin{table*}[ht]
\centering
\begin{tabular}{l|cc|ccc|c}
\toprule
            &   \multicolumn{2}{c}{\bf Human}  &    \multicolumn{3}{c}{\bf Baselines}     & \multicolumn{1}{c}{\bf Ours} \\
\bf Game        & \bf Max & \bf Walkthrough-100   & \bf TDQN  & \bf DRRN  & \bf KG-A2C & \bf \ourshort  \\ 
\midrule
\textcolor{EasyColor}{detective}   & 360 & 350  & 169   & 197.8 & 207.9 & 	\textbf{317.7} \\ 
\textcolor{HardColor}{enchanter}   & 400 & 125  & 8.6   & 	{20}    & 12.1  & 	\textbf{21.6}  \\ 
\textcolor{EasyColor}{inhumane}    & 90 & 70   & 0.7   & 0     & 	{3}     & \textbf{15.83}  \\ 
\textcolor{MediumColor}{karn}        & 170 & 40  & 0.7   & \textbf{2.1}   & 0     & 	{0.0}  \\ 
\textcolor{EasyColor}{snacktime}   & 50 & 50   &  9.7   & 0     & 0     & \textbf{20}\\ 
\textcolor{HardColor}{spellbrkr}   & 600 & 160  & 18.7  & 	{37.8}  & 21.3  & \textbf{40}\\ %
\textcolor{MediumColor}{zork1}       & 350 & 102  & 9.9   & 32.6  & 34    & \textbf{43.58}\\
\textcolor{MediumColor}{zork3}       & 7  & 3   & 0     & 0.5   & 0.1   & 	\textbf{2.67}\\
\bottomrule
\end{tabular}
\caption{\label{tab:jericho_results_subset}{Maximum scores on a subset of Jericho games (selected randomly based on the difficulty level) achieved by the agents (proposed and baseline) averaged over $10$ runs. Difficulty levels: easy marked in \textcolor{EasyColor}{green} color, difficult in \textcolor{MediumColor}{tan}, and extreme in \textcolor{HardColor}{red}.}}
\end{table*}

Next, we consider \jericho~\cite{hausknecht19}, a benchmark dataset in TBGs that consists of $33$ popular interactive fiction (IF) games developed for humans a decade ago. We randomly select a subset of games from different difficulty levels for our experiments.
From Table~\ref{tab:jericho_results_subset}, we can see that \ourshort\ outperforms the other state-of-the-art text-only baselines (Template DQN \cite{hausknecht19}, DRRN \cite{narasimhan2015language, he2016deep}, and KG-A2C \cite{ammanabrolu2019graph}) by a significant margin. Our approach is currently able to achieve the best score (averaged over $10$ runs) on $7/8$ games from across the difficulty levels.

\subsubsection{Images: Retrieval vs. Generation}

\begin{figure*}[htbp]
    \centering
    \includegraphics[width=1.0\textwidth]{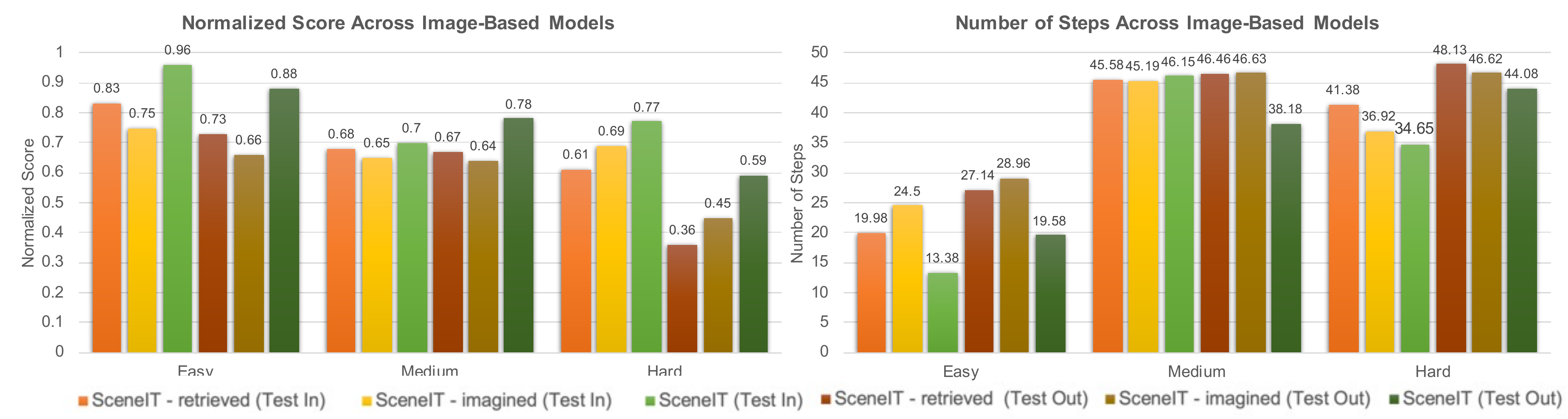}
    \caption{Results showing an improvement \textcolor{black}{across both normalized score (higher is better) and number of steps (lower is better)} by using images on the TWC dataset with different difficulty levels.}
    \label{fig:results_image_models_improvement}
\end{figure*}

After establishing that the addition of the visual features from images that represent the scene described by the textual observations from the game does indeed help the performance of agents, we now explore further into comparison between these different agents. Specifically, we compare the three models described in Section~\ref{sec:models}: \ourshort\ with retrieved images from the internet, \ourshort\ with generated/imagined images from the pretrained AttnGAN and \ourshort\ with finetuned AttnGAN. This comparison is presented as a bar chart in Figure~\ref{fig:results_image_models_improvement}. As in the previous experiments, we plot the three difficulty levels across two conditions: {\tt in} and {\tt out} of distribution. We use a lighter shade of the corresponding color for the former, and a darker shade for the latter. It is clear that \ourshort\ -- which combines text features with features from AttnGAN -- outperforms the other two image baselines across different difficulty levels and conditions.

\subsection{Qualitative Results}
\label{subsec:qualitative_results}

\begin{figure*}[t]
		\centering
		\includegraphics[width=1.0\linewidth]{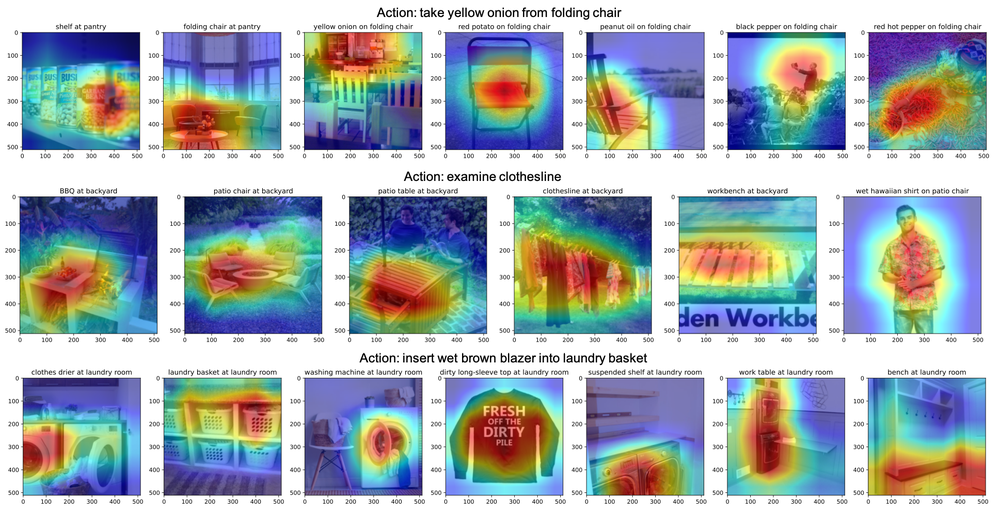}
		\caption{Activation maps showing the region of interest when producing the action command in each case, using the internet-based retrieval model for \twc.}
		\label{fig:google_att}
\end{figure*}

\begin{figure*}[h]
		\centering
		\includegraphics[width=1.0\linewidth]{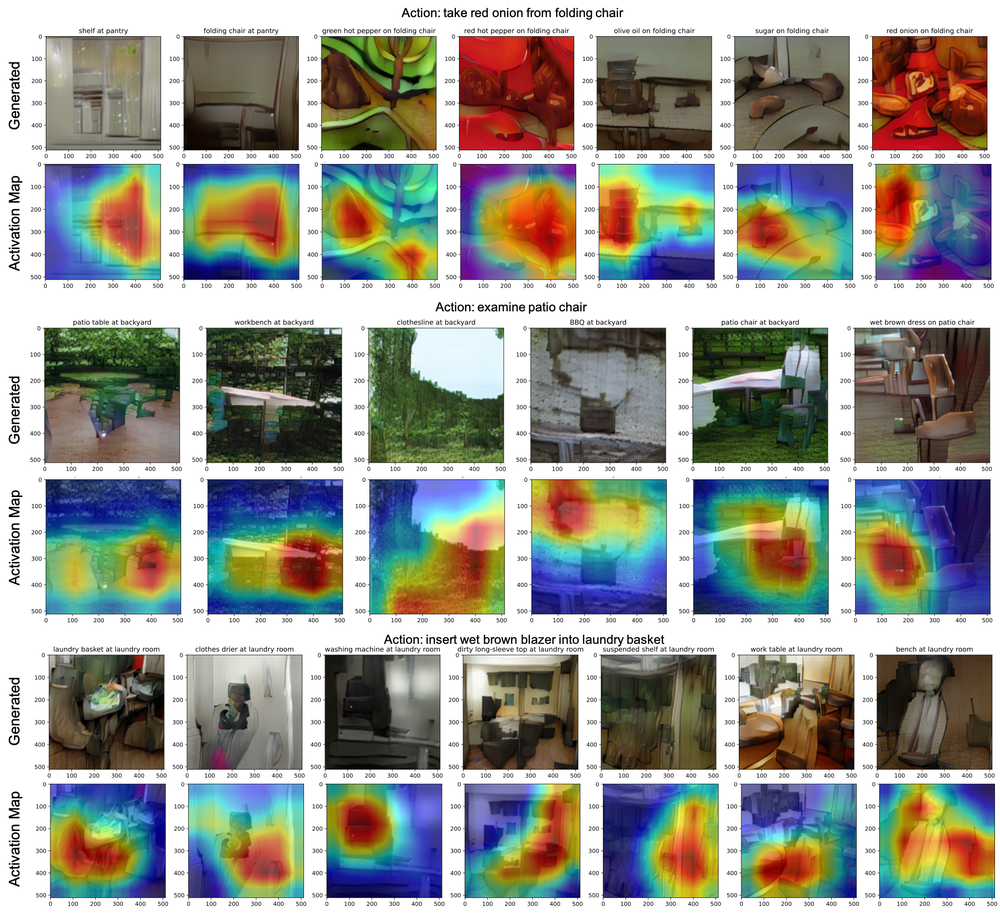}
		\caption{Activation maps showing the region of interest when producing the action command in each case, using the imagination based model for \twc. We include both the generated images and its attention plot for clarity.}
		\label{fig:attngan_att}
\end{figure*}

In addition to the quantitative results described previously, we also present some qualitative examples of what the \ourshort\ agent focuses on as it uses images (retrieved or imagined) in order to solve specific problem instances. To illustrate this effectively, we use the notion of attention activation maps~\cite{zhou2016learning,selvaraju2017grad,lu2012learning,gupta2021recognition}, which can be used to demonstrate parts of an image that an agent/technique is attending to. We split our analysis into the two main ways in which we currently produce images for use by \ourshort: retrieval, and imagination (see Section~\ref{sec:methodology}). 

Figure~\ref{fig:google_att} shows examples of this for images retrieved from the Internet. We present three examples of the various images that are produced for a given text phrase from the observation as input (e.g. {\tt clothesline at backyard}), as well as the final action that is taken by the agent (e.g. {\tt examine clothesline}). The other two examples follow a similar pattern; and together, these three examples illustrate that \ourshort\ can focus the agent's attention on relevant parts of the retrieved image to facilitate the final decision making. 

A similar pattern is seen in the case of imagined images: Figure~\ref{fig:attngan_att} presents both the imagined images as well as the activation maps overlaid over those respective images for a given set of text phrases from the game observation. For example, the agent can focus on the right part of the image that is imagined for the phrase {\tt wet brown dress on patio chair}; and can then choose the action {\tt examine patio chair}. The other examples also illustrate a similar pattern.

This analysis also presents an interesting contrast between images that are retrieved from the Internet, versus ones that are generated from scratch by an AI model. For example, consider the two text phrases {\tt patio chair at backyard} and {\tt clothesline at backyard}. Both these phrases and the images retrieved and generated respectively for them appear in both Figure~\ref{fig:google_att} and Figure~\ref{fig:attngan_att} -- however, the visual representation of the pairs of pictures 
is strikingly different. The \ourshort\ agent is also led to choose different actions in the two cases -- while in the case of the retrieval it chooses to examine the clothesline, in the case of imagination it instead chooses to examine the patio chair. These examples illustrate qualitatively how the two image retrieval techniques can work in different and often complementary ways.

\section{Related Work}
\label{sec:related_work}

The field of text-based and interactive games has seen a lot of recent interest and work, thanks in large part to the creation and availability of pioneering environments such as \tw~\cite{cote2018textworld} and the \jericho~\cite{hausknecht19} collection. Based on these domains, several interesting approaches have been proposed that seek to improve the efficiency of agents in these environments~\cite{ammanabrolu2019playing,dambekodi2020playing,chaudhury2020bootstrapped,murugesan2021text}. We mention and discuss this prior work in context in the earlier parts of this paper.

Separate from this progress on TBGs, there has also been work on Inductive Logic Programming (ILP) methods -- these methods have shown good relation generalization in symbolic domains using differentiable model learning on symbolic inputs~\cite{evans2018learning, richardson2006markov}, even in noisy settings. Neural Logic Machines~\cite{dong2019neural} have shown good generalization to out-of-sample games using dedicated MLP units for first-order rule learning by interacting with the environment. The work on Logical Neural Networks~\cite{riegel2020logical} is a recent addition to the family of ILP methods that can learn differentiable logical connectives using constrained optimization over the differentiable neural network framework. Concurrently, there has been work in the (symbolic) automated planning community that has looked at learning and inferring the relations (predicates) that make up an underlying domain -- like the eight-tile puzzle -- by using variational auto-encoders~\cite{asai2019unsupervised,asai2018classical,asai2020learning,asai2020discrete}.

\section{Conclusion}
\label{sec:conclusion}

In this paper, we introduced \ourlong\ (\ourshort), a model for RL agents executing in text-based games. \ourshort\ uses the text from observations provided by the game to either retrieve or generate images that correspond to the scene represented by the text; and then combines the features from the images along with features from the text in order to select the next best action for the RL agent. We show via an extensive experimental evaluation that \ourshort\ shows better performance -- in terms of the normalized reward score achieved by agents, as well as the number of steps to complete a task -- than existing state-of-the-art models that rely only on the observation text. We also presented qualitative results that showed that an agent guided by \ourshort\ focuses its attention on those parts of an image that we may expect a human to attend to as well.

\newpage

\bibliographystyle{plain}
\bibliography{references.bib}

\newpage
\appendix
\section{Experimental Details}

In this section, we report the experimental setup and settings used in our paper.

\textbf{TBGs as a POMDP:} TBGs can be framed as partially observable Markov decision processes (POMDPs)~\cite{spaan2012partially} denoted $ \langle S, A, O, T, E, r \rangle $, where:
$S$ denotes the set of states, $A$ denotes the action space, $O$ denotes the observation space, $T$ denotes the state transition probabilities, $E$ denotes the conditional observation emission probabilities, and $r: S \times A \rightarrow \mathbb{R}$ is the reward function.
The observation $o_t$ at time step $t$ depends on the current state. Both observations and actions are rendered in text.
The agent receives a reward at every time step
$t$: $r_t=r(o_t,a_t)$, and the agent's goal is to maximize the %
expected discounted sum of rewards:
$\mathbb{E}[\sum_t\gamma^{t} r_t]$, where $\gamma \in [0, 1]$ is a discount factor. In our experiments, we set $\gamma = 0.9$.  All policies are learned using \textit{Actor-Critic} \cite{mnih2016asynchronous}. We use \textit{GloVe} word embeddings to represent our observation text and project it to a $64$ dimensional vector. We use $128$ dimensions as our hidden size in the (bidirectional) text encoder. 

\textbf{\twclong}\ consists of 30 games per difficulty level (listed below) that have been generated and separated into 3 directories: train, test (OUT), and valid (IN). The goal of these games is \textit{house cleanup}, where each object is misplaced in a house (with one or more rooms) and the agent needs to return the misplaced objects to a commonsensically appropriate location (e.g., apple to the refrigerator and dirty sock to the laundry basket, etc). The objects in these games follow commonsensical relations and previous work \cite{murugesan2021text} has shown that leveraging external commonsense knowledge (such as \texttt{ConceptNet}) improves the performance.

\begin{itemize}
    \item Easy level games have only 1 room and up to 3 objects that need to be placed in their appropriate location.
    \item Medium level games have 1 room with up to 5 objects.
    \item Hard level games have either 4 or 5 objects shuffled across 3 or 4 rooms.
\end{itemize}

In our experiments with \twc, we set the maximum number of episodes to $100$ with a $50$ step maximum per episode.

\textbf{\textit{First TextWorld Problems}}\ consist of $4,440$ different training games, $222$ validation games (IN), and $514$ test games (OUT). These belong to the cooking domain, where the cooking ingredients are placed throughout the house and the agent has to collect these objects/items (listed in the cookbook) to prepare a delicious meal. Unlike in \twc, the objects in these games are not related and the agent needs to utilize its exploration capabilities to collect these items. 

In our experiments with \ftwp, we set the maximum number of episodes to $5$ with $100$ steps per episode, as suggested in prior work \cite{adolphs2020ledeepchef}. 

In \textbf{\jericho}\, the agent can take up to $100$ steps per episode with a limit of $100,000$ maximum total steps per run. The Jericho environment allows handicap configurations for the agents when interacting with the environment. Depending on the handicap level, the agent may request additional information from the environment for better exploration.  We follow prior work (baselines reported in Table 2 in the main paper) for our handicap configuration for fair comparison among the baselines. Specifically, we generate a set of valid actions per step using the action templates and evaluate it using the \jericho\ environment to check whether the action is valid. In addition to the observation text, we use the additional information: \textit{inventory} and \textit{look} from the \jericho\ environment. Please see Section \ref{sec:gameplay} for a sample gameplay readout for \textit{Zork1} -- a benchmark text-based interactive fiction game developed in the 1980s which is a part of the \jericho\ game suite.

\textbf{Resources:} The agents were trained in parallel on two machines with the following specifications:

\begin{table}[!h]
\centering
{%
\begin{tabular}{l|l}
\textbf{Resource} & \textbf{Setting}                                   \\  \hline 
CPU      & Intel(R) Xeon(R) CPU E5-2690 v4 @ 2.60GHz \\
Memory   & 128GB                                     \\
GPUs     & 2 x NVIDIA Tesla V100 16 GB               \\
Disk1    & 100GB                                     \\
Disk2    & 600GB                                     \\
OS       & Ubuntu 18.04-64 Minimal for VSI.         
\end{tabular}
}
\vspace{5mm}
\caption{Resources used by the agents.}
\end{table}
Each agent was trained on a single GPU for approximately 12 hours for the \textit{Text} agent and 16 hours for the \ourshort\ agent for each run. We use \textit{SpaCy} with additional hand-written rules to extract the noun phrases for the image retrieval/generation.

\section{Additional Experimental Results}

\subsection{Quantitative Results: Training Curves for Different \ourshort\ Strategies}

Figure \ref{fig:res_twc} shows the training curves for different strategies for utilizing images in our proposed \ourshort\ approach. We compare the agent's performance based on both retrieved and generated images, to complement the test set (IN and OUT distributions) performance results reported in Section 3.1.4 of the main paper. We use \textit{Text + Google} (\ourshort\ retrieved) for the internet retrieved images and \textit{Text + AttnGAN} (\ourshort\ imagined/generated) for AttnGAN generated images. \textit{Text + ModelGAN} (\ourshort) shows the proposed \ourshort\ method that fine-tunes the pretrained AttnGAN while training our agent -- this agent is used for all our experiments, as mentioned in Section 3 of the main paper. The training curves show that both \textit{Text + Google} and \textit{Text + AttnGAN} feature improvements in the early episodes of the training; whereas \textit{Text + ModelGAN} slowly catches up to the other approaches. This is due to the fine-tuning of the pre-trained AttnGAN along with the training of the agent. The test performance of \ourshort\ with the fine-tuning approach validates that this approach outperforms other strategies and the baselines.

\subsection{Qualitative Results: Activation Maps for \ftwp\ and \jericho\ }

\begin{figure*}[h]
	\centering
	\includegraphics[width=1.0\linewidth]{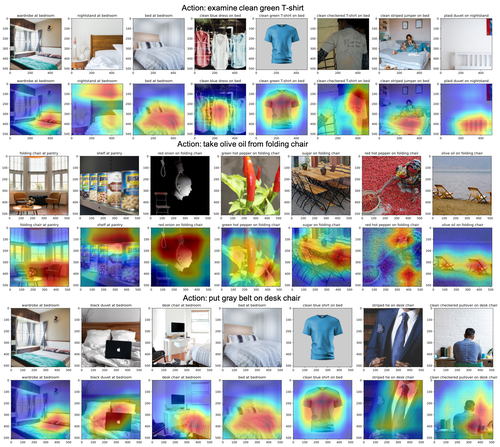}
	\caption{Activation maps for \textit{First TextWorld Problems} Cooking Task when using images retrieved from internet for selecting the next action. We show the retrieved images along with its attention plot for clarity.}
	\label{fig:google_ftwp}
\end{figure*}

\begin{figure*}[h]
	\centering
	\includegraphics[width=1.0\linewidth]{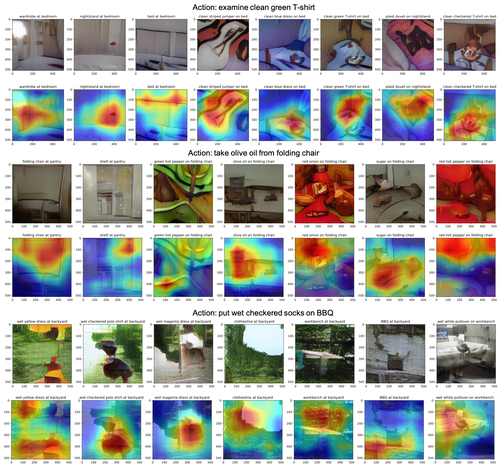}
	\caption{Activation maps for \textit{First TextWorld Problems} Cooking Task showing the region of interest when using the imagination based model (AttnGAN) for selecting the next action. We include both the generated images and its attention plot for clarity.}
	\label{fig:attngan_ftwp}
\end{figure*}

\begin{figure*}[h]
	\centering
	\includegraphics[width=1.0\linewidth]{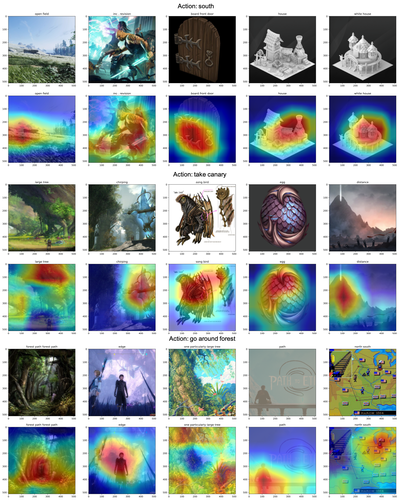}
	\caption{Activation maps for \textit{Zork1} from Jericho environment, when using images retrieved from the internet for selecting the next action. We show the retrieved images along with its attention plot for clarity.}
	\label{fig:zork1_google}
\end{figure*}

\begin{figure*}[h]
	\centering
	\includegraphics[width=1.0\linewidth]{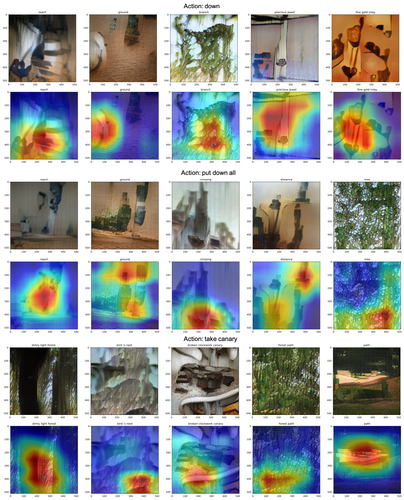}
	\caption{Activation maps for \textit{Zork1} from Jericho environment,  on the imagined images from Zork1 games using AttnGAN by the visual model. }
	\label{fig:zork1_attngan}
\end{figure*}

Figures \ref{fig:google_ftwp} and \ref{fig:attngan_ftwp} shows the activation maps for the \textit{First TextWorld Problems}. Figures \ref{fig:zork1_google} and \ref{fig:zork1_attngan} show the activation maps for \jericho. The title on each sub-figure shows the next action taken based on the observation text, and the images generated/retrieved using the objects and the relational phrases between the objects in the scene from the observation.

From each observation at time $t$, we extract various keywords that represent the relationships between the objects in the scene. As shown in Figures \ref{fig:attngan_ftwp} and \ref{fig:google_ftwp}, for each keyword, we either retrieve the image from the web or generate using AttnGAN. Our model uses the various images for each keyword to generate the action string using a combination of textual and visual features. We use the ResNet-18 model for the visual feature extraction, which we fine-tune during the training of the agent. During inference, we use GradCAM~\cite{selvaraju2017grad} on the visual model to extract regions of interest in the image that is used for feature extraction; this is then ultimately used for generating the final action (shown as the heading for each image) from the list of admissible actions.

Note that in the case of the \jericho\ environment, the AttnGAN-based generative model encounters out-of-sample query strings (i.e., phrases extracted from observation text) which are unique to these fantasy games such as ``precious jewel'', ``broken clockwork canary'' etc, that are not seen during training.

\section{Sample Gameplay by the Proposed \ourshort\ Agent for \textit{Zork1} from \jericho\ }

In this section, we show a sample of gameplay by our agent in \textit{Zork1}, a benchmark text-based interactive fiction game developed in the 1980s (now a part of the \jericho\ game suite). We show the generated images from our \ourshort\ model with fine-tuned AttnGAN for some of the steps for clarity.

\label{sec:gameplay}
\begin{verbatim}
[Action Taken]:look
[Reward]:0
[Score]:0
**Step 95078**
[Obs]:Copyright (c) 1981, 1982, 1983 Infocom, Inc. All rights reserved.
ZORK is a registered trademark of Infocom, Inc.
Revision 88 / Serial number 840726
West of House
You are standing in an open field west of a white house, with a boarded front door.
There is a small mailbox here.
\end{verbatim}
\begin{figure*}[h]
    \centering
        \includegraphics[scale=0.4]{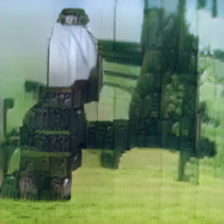} \includegraphics[scale=0.4]{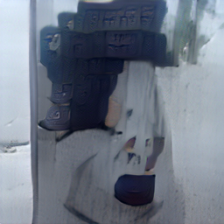}
        \includegraphics[scale=0.4]{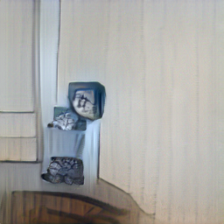}
        \includegraphics[scale=0.4]{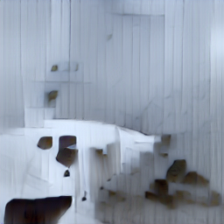}
    \caption{\ourshort\ generated images for this observation text.}
    \label{fig:res_twc}
\end{figure*}
\begin{verbatim}
[Action Taken]:open mailbox
[Reward]:0
[Score]:0
**Step 95079**
[Obs]:opening the small mailbox reveals a leaflet.
[Action Taken]:west
[Reward]:0
[Score]:0
**Step 95080**
[Obs]:forest
this is a forest, with trees in all directions. to the east, there appears to be sunlight.
[Action Taken]:go around trees
[Reward]:0
[Score]:0
**Step 95081**
[Obs]:forest
this is a dimly lit forest, with large trees all around.
[Action Taken]:go around trees
[Reward]:0
[Score]:0
**Step 95082**
[Obs]:forest
this is a dimly lit forest, with large trees all around.
[Action Taken]:west
[Reward]:0
[Score]:0
**Step 95083**
[Obs]:forest
you hear in the distance the chirping of a song bird.
[Action Taken]:south
[Reward]:0
[Score]:0
**Step 95084**
[Obs]:forest
[Action Taken]:north
[Reward]:0
[Score]:0
**Step 95085**
[Obs]:clearing
you are in a small clearing in a well marked forest path that extends to the east and west.
[Action Taken]:go around forest
[Reward]:0
[Score]:0
**Step 95086**
[Obs]:forest
[Action Taken]:north
[Reward]:0
[Score]:0
**Step 95087**
[Obs]:clearing
you are in a clearing, with a forest surrounding you on all sides. 
a path leads south.
on the ground is a pile of leaves.
[Action Taken]:take pile
[Reward]:0
[Score]:0
**Step 95088**
[Obs]:in disturbing the pile of leaves, a grating is revealed.
taken.
[Action Taken]:put down pile
[Reward]:0
[Score]:0
**Step 95089**
[Obs]:dropped.
[Action Taken]:south
[Reward]:0
[Score]:0
**Step 95090**
[Obs]:forest path
this is a path winding through a dimly lit forest. the path heads 
north-south here. one particularly large tree with some low branches 
stands at the edge of the path.
you hear in the distance the chirping of a song bird.
[Action Taken]:up
[Reward]:0
[Score]:0
**Step 95091**
[Obs]:up a tree
you are about 10 feet above the ground nestled 
among some large branches. the nearest branch above you is above your reach.
beside you on the branch is a small bird's nest.
in the bird's nest is a large egg encrusted with 
precious jewels, apparently scavenged by a childless songbird. 
the egg is covered with fine gold inlay, and ornamented in 
lapis lazuli and mother-of-pearl. unlike most eggs, 
this one is hinged and closed with a delicate looking clasp. 
the egg appears extremely fragile.
\end{verbatim}
\begin{figure*}[h]
    \centering
        \includegraphics[scale=0.4]{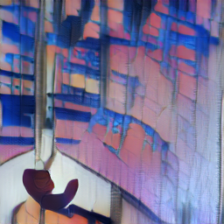} \includegraphics[scale=0.4]{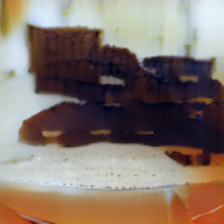}
        \includegraphics[scale=0.4]{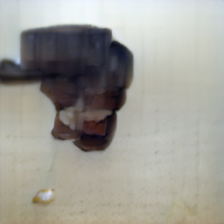}
        \includegraphics[scale=0.4]{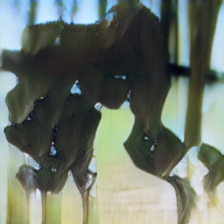}
    \caption{\ourshort\ generated images for this observation text.}
    \label{fig:res_twc}
\end{figure*}
\begin{verbatim}
[Action Taken]:down
[Reward]:0
[Score]:0
**Step 95092**
[Obs]:forest path
[Action Taken]:south
[Reward]:0
[Score]:0
**Step 95093**
[Obs]:north of house
you are facing the north side of a white house. 
there is no door here, and all the windows are boarded up. 
to the north a narrow path winds through the trees.
[Action Taken]:west
[Reward]:0
[Score]:0
**Step 95094**
[Obs]:west of house
there is a small mailbox here.
the small mailbox contains:
  a leaflet
[Action Taken]:north
[Reward]:0
[Score]:0
**Step 95095**
[Obs]:north of house
[Action Taken]:north
[Reward]:0
[Score]:0
**Step 95096**
[Obs]:forest path
[Action Taken]:go around forest
[Reward]:0
[Score]:0
**Step 95097**
[Obs]:clearing
[Action Taken]:west
[Reward]:0
[Score]:0
**Step 95098**
[Obs]:behind house
you are behind the white house. a path leads into the 
forest to the east. in one corner of the house there is 
a small window which is slightly ajar.
[Action Taken]:south
[Reward]:0
[Score]:0
**Step 95099**
[Obs]:south of house
you are facing the south side of a white house. 
there is no door here, and all the windows are boarded.
\end{verbatim}
\begin{figure*}[h]
    \centering
        \includegraphics[scale=0.4]{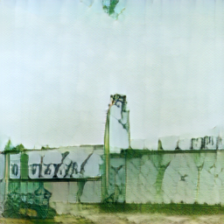} \includegraphics[scale=0.4]{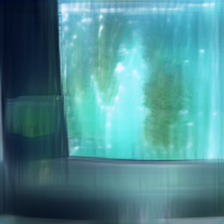}
        \includegraphics[scale=0.4]{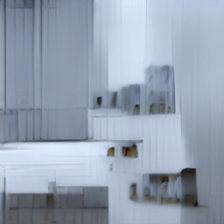}
        \includegraphics[scale=0.4]{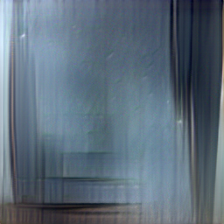}
    \caption{\ourshort\ generated images for this observation text.}
    \label{fig:res_twc}
\end{figure*}
\begin{verbatim}
[Action Taken]:east
[Reward]:0
[Score]:0
**Step 95100**
[Obs]:behind house
[Action Taken]:open window
[Reward]:0
[Score]:0
**Step 95101**
[Obs]:with great effort, you open the window far enough to allow entry.
[Action Taken]:west
[Reward]:10
[Score]:10
**Step 95102**
[Obs]:kitchen
you are in the kitchen of the white house. 
a table seems to have been used recently for the preparation of food. 
a passage leads to the west and a dark staircase can be seen leading upward. 
a dark chimney leads down and to the east is a small window 
which is open.on the table is an elongated brown sack, 
smelling of hot peppers.
a bottle is sitting on the table.
the glass bottle contains:
  a quantity of water
  \end{verbatim}
\begin{figure*}[h]
    \centering
        \includegraphics[scale=0.4]{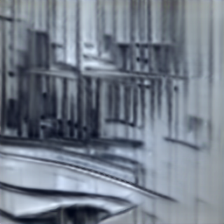}
        \includegraphics[scale=0.4]{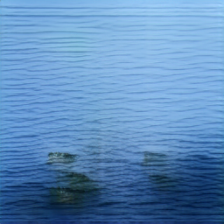}
        \includegraphics[scale=0.4]{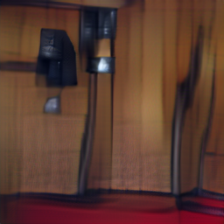}
        \includegraphics[scale=0.4]{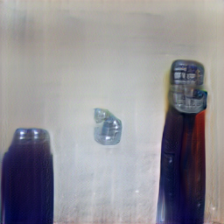}
    \caption{\ourshort\ generated images for this observation text.}
    \label{fig:res_twc}
\end{figure*}
\begin{verbatim}
[Action Taken]:west
[Reward]:0
[Score]:10
\end{verbatim}

\begin{figure*}[htbp]
    \centering
        \includegraphics[trim={0.45cm 0.0cm 0.25cm 1.5cm},scale=0.445,clip]{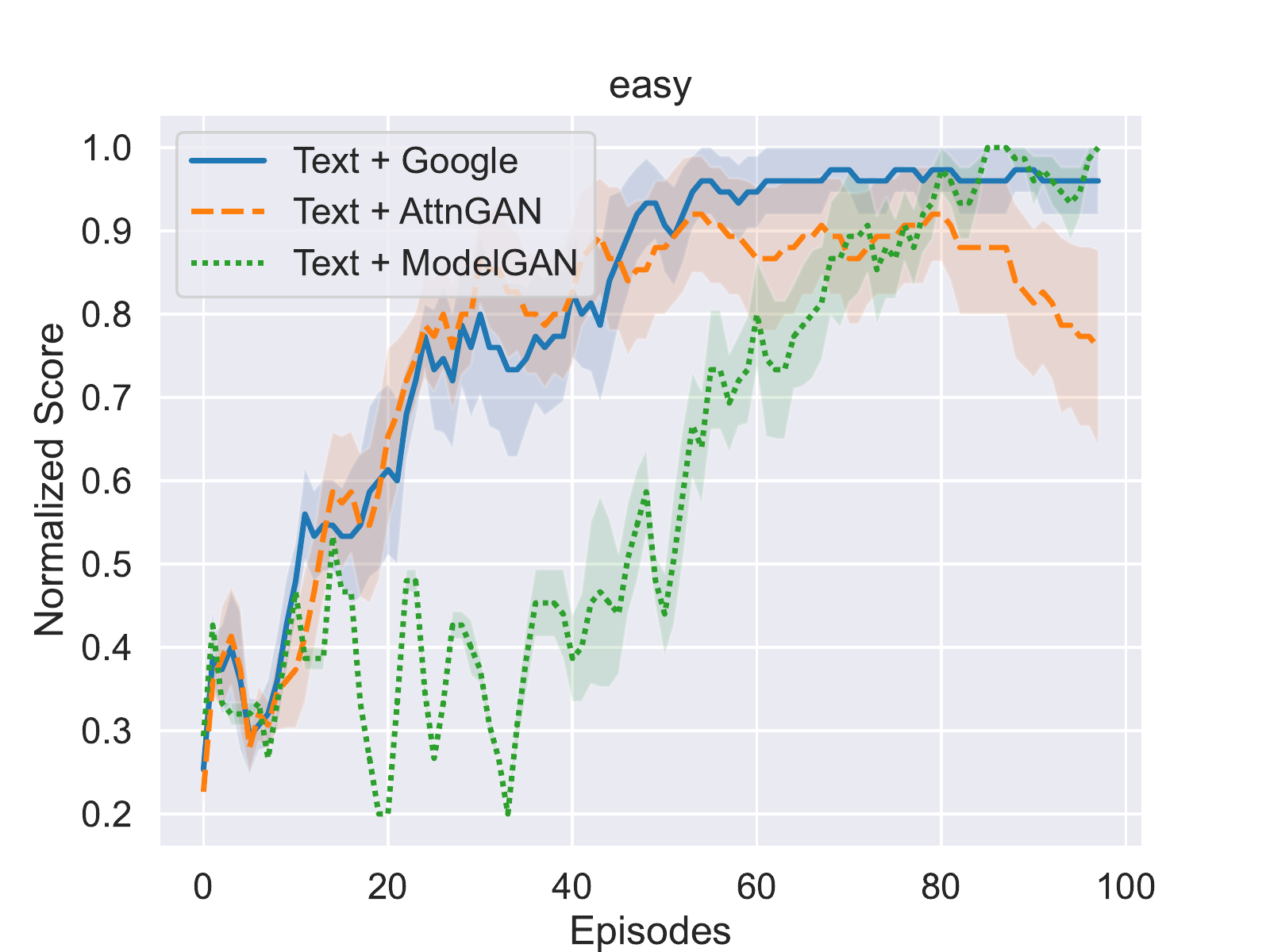} 
        \includegraphics[trim={0.6cm 0.0cm 0.25cm 1.5cm},scale=0.445,clip]{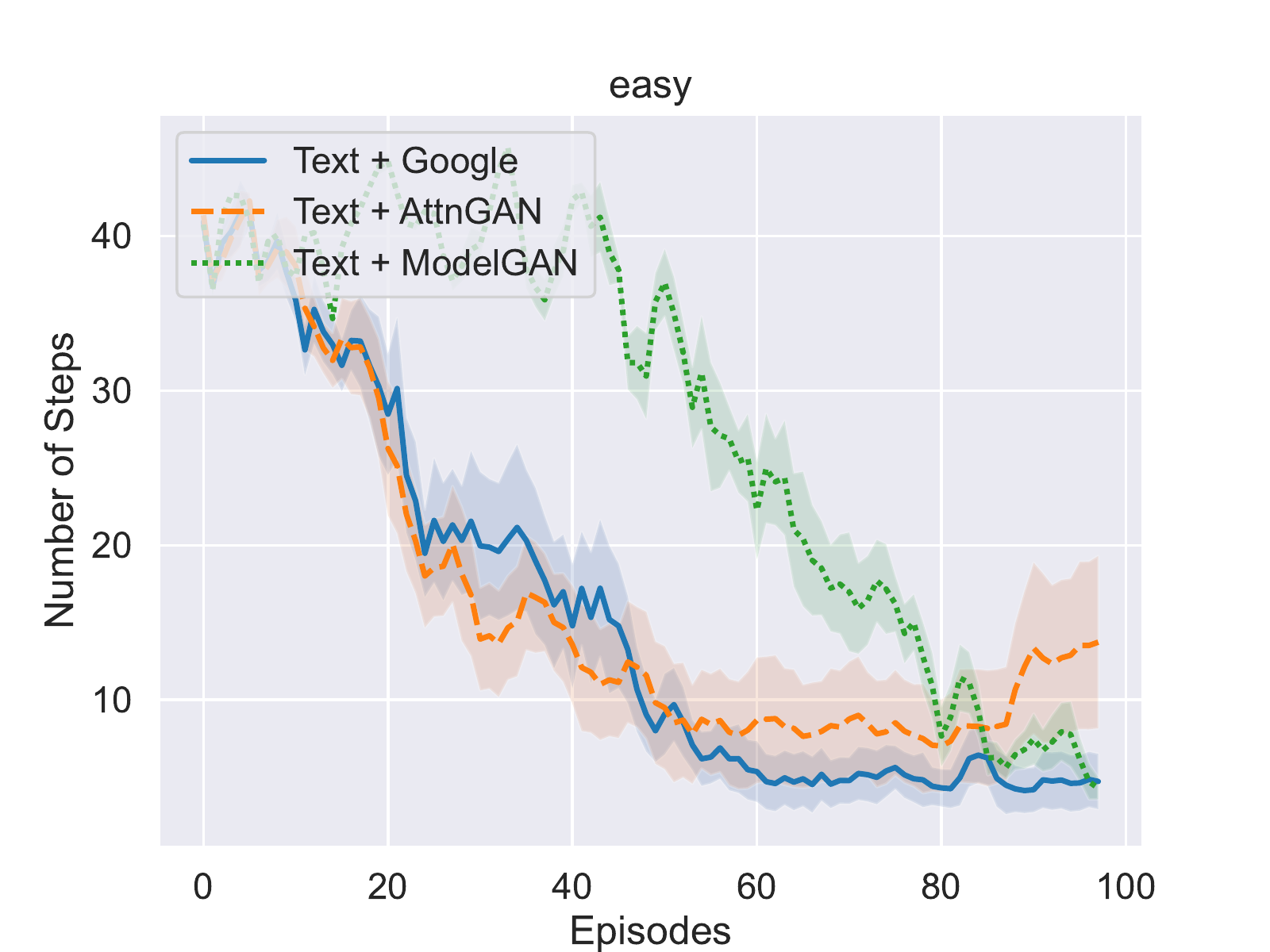}\\
        \includegraphics[trim={0.45cm 0.0cm 0.25cm 1.5cm},scale=0.445,clip]{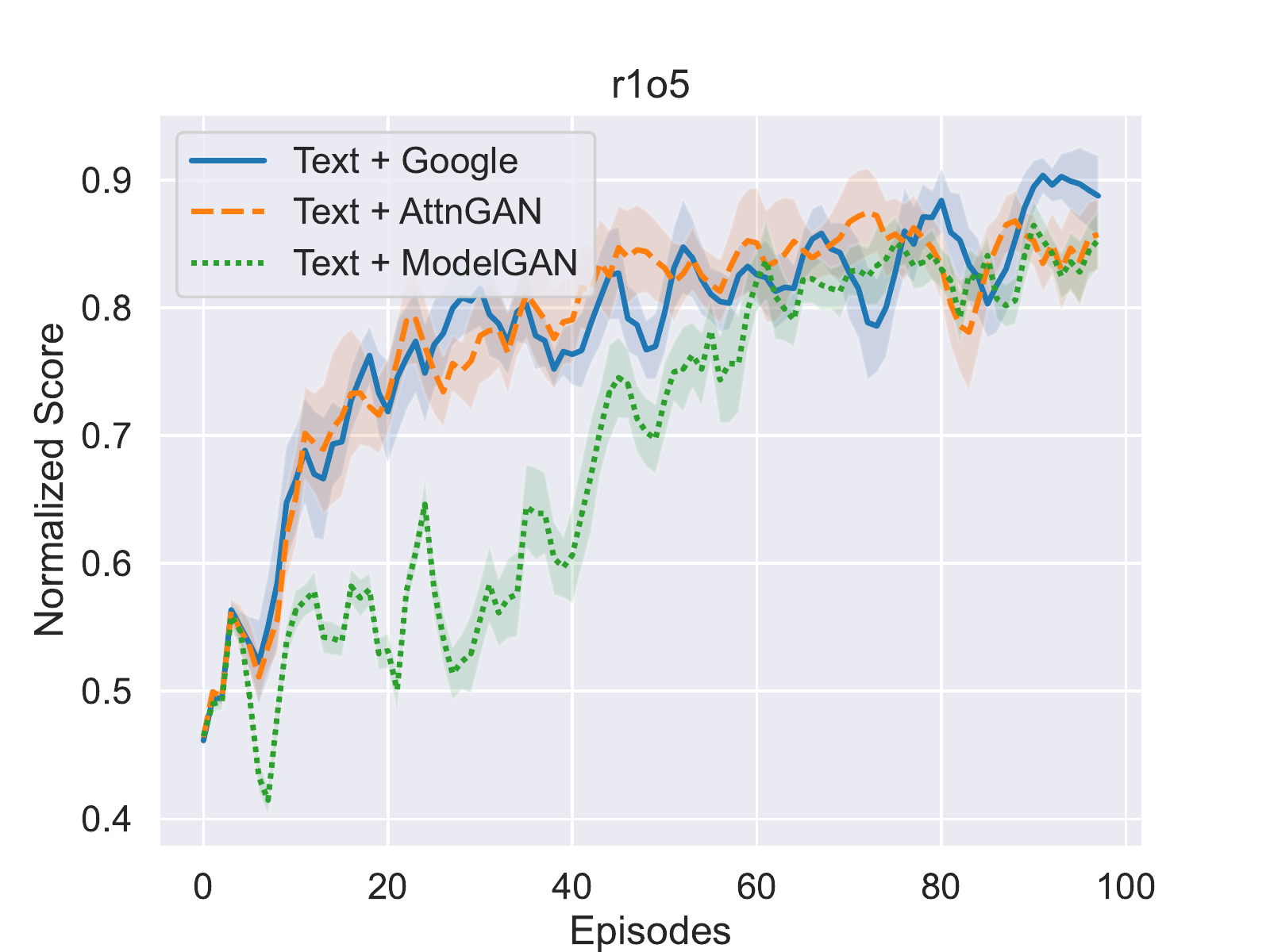}
        \includegraphics[trim={0.6cm 0.0cm 0.25cm 1.5cm},scale=0.445,clip]{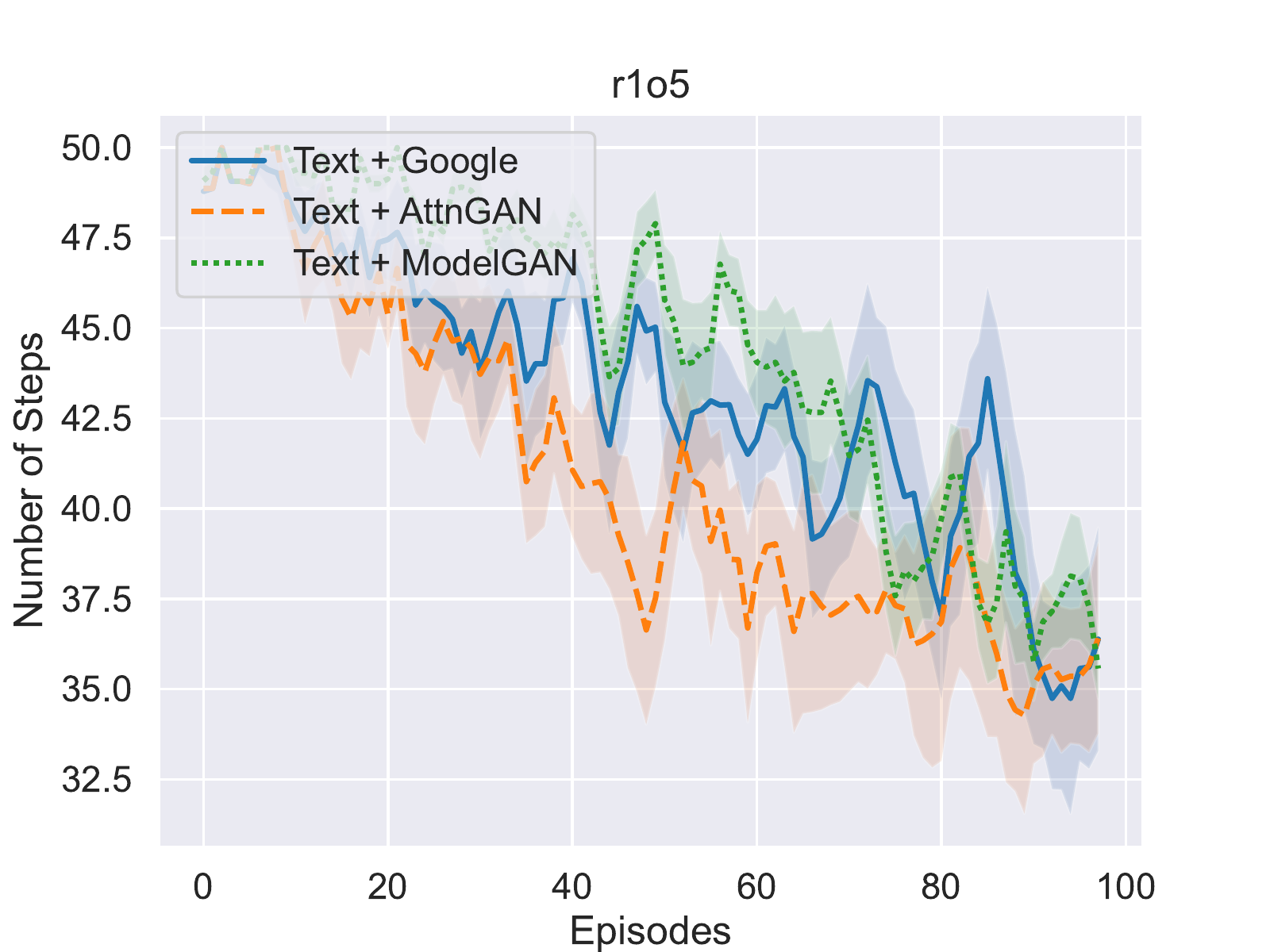}\\
        \includegraphics[trim={0.45cm 0.0cm 0.25cm 1.5cm},scale=0.445,clip]{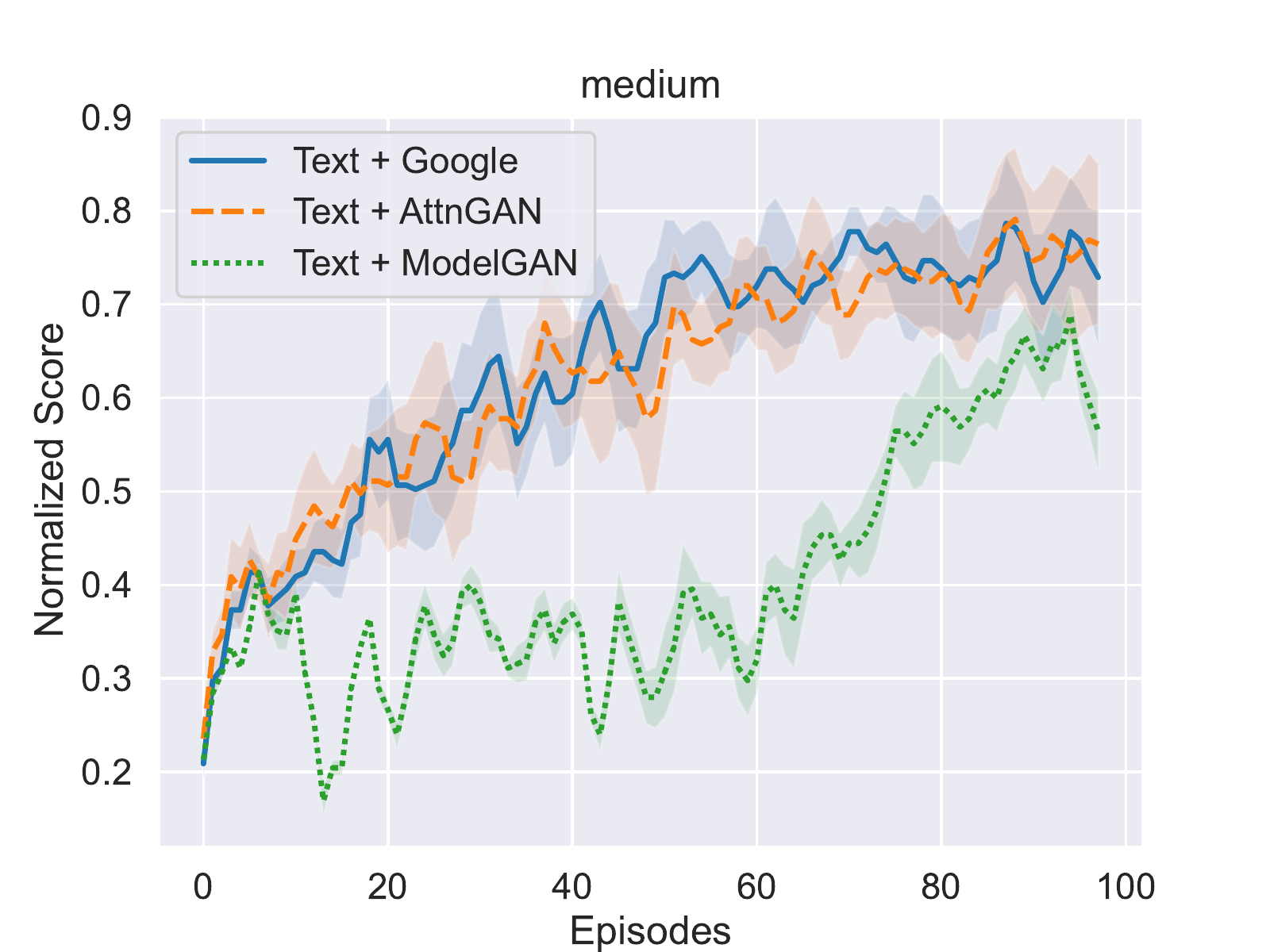}
        \includegraphics[trim={0.6cm 0.0cm 0.25cm 1.5cm},scale=0.445,clip]{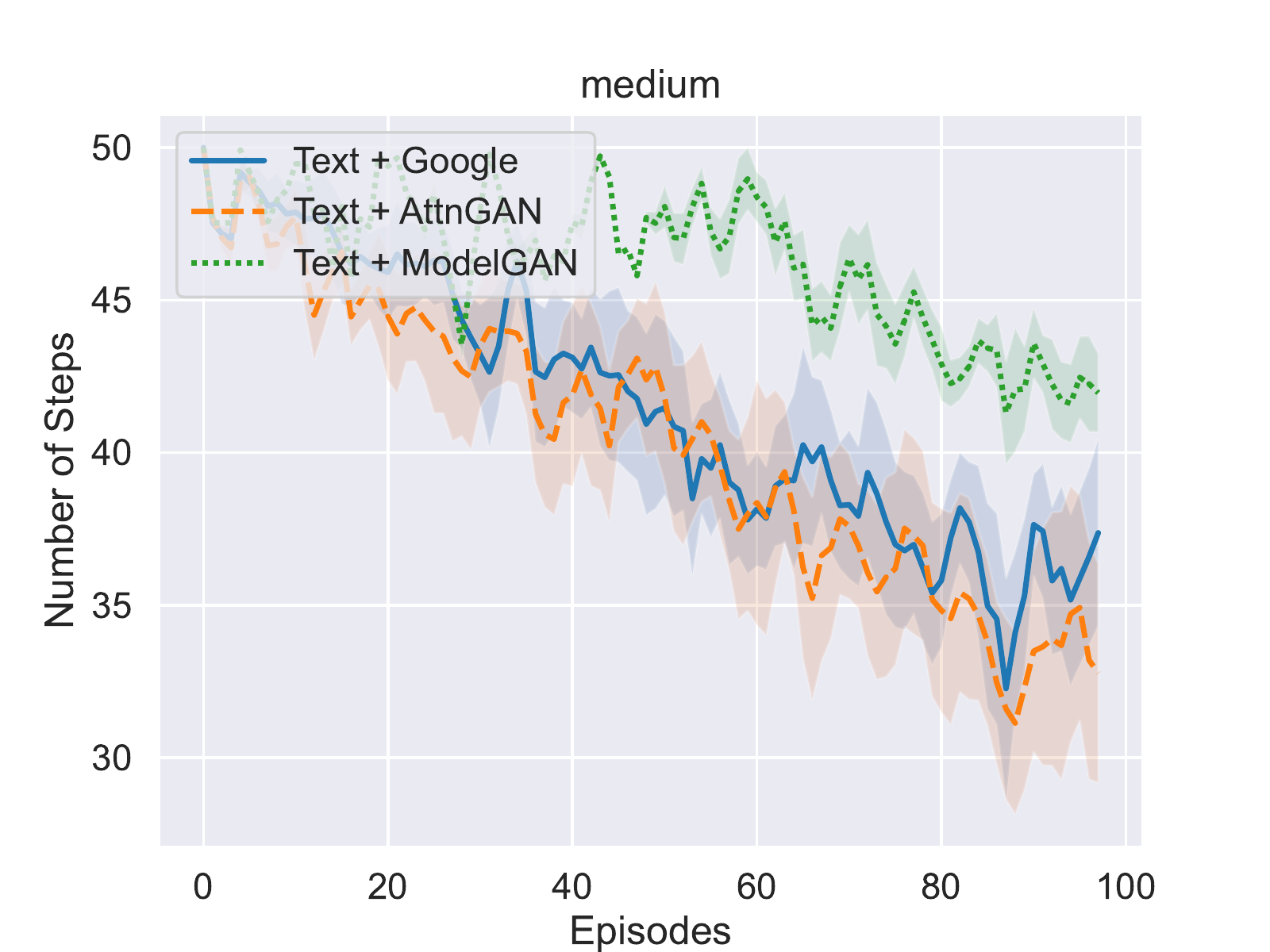}
    \caption{Performance evaluation on TextWorld Commonsense (TWC) across the three visual methods measured during training (showing mean and standard deviation averaged over $5$ runs) for the three difficulty levels: Easy (Top), Medium (Middle), Hard (Bottom) using normalized score and the number of steps taken.}
    \label{fig:res_twc}
\end{figure*}

\end{document}